\documentclass{article}

     \PassOptionsToPackage{numbers, compress}{natbib}

\usepackage[preprint]{neurips_2023}

\usepackage[utf8]{inputenc} 
\usepackage[T1]{fontenc}    
\usepackage{hyperref}      
\usepackage{url}            
\usepackage{booktabs}       
\usepackage{amsfonts}       
\usepackage{nicefrac}       
\usepackage{microtype}      
\usepackage{xcolor}         
\hypersetup{
    colorlinks=true,
    linkcolor=blue,
    filecolor=magenta,      
    urlcolor=cyan,
    pdftitle=GRAtt-VIS,
    pdfpagemode=FullScreen,
    }
\usepackage{graphicx}
\usepackage[export]{adjustbox}
\usepackage{multirow}
\usepackage{pifont}
\usepackage{amssymb}
\usepackage{amstext}
\usepackage{amsmath}
\usepackage[bb=boondox]{mathalfa}
\newcommand{\cmark}{\ding{51}}%
\newcommand{\name}{GRAtt-VIS}
\usepackage{natbib}
\usepackage{float}
\title{ \name{}: Gated Residual Attention for Auto Rectifying Video Instance Segmentation }

\author{
Tanveer Hannan$^{1,2\thanks{Joint first-authorship and equal contributions.}}$, Rajat Koner$^{1,2^*}$, Maximilian Bernhard$^{1,2}$, Suprosanna Shit$^3$,\\ 
\textbf{Bjoern Menze}$^4$, \textbf{Volker Tresp}$^{1,2}$, \textbf{Matthias Schubert}$^{1,2}$, \textbf{Thomas Seidl}$^{1,2}$
 \\
$^1$ LMU Munich, $^2$ MCML, $^3$ Technical University of Munich, $^4$ University of Zurich\\
\{hannan, koner\}@dbs.ifi.lmu.de
}

\begin{document}

\maketitle

\begin{abstract}
Recent trends in Video Instance Segmentation (VIS) have seen a growing reliance on online methods to model complex and lengthy video sequences. However, the degradation of representation and noise accumulation of the online methods, especially during occlusion and abrupt changes, pose substantial challenges. Transformer-based query propagation provides promising directions at the cost of quadratic memory attention. However, they are susceptible to the degradation of instance features due to the above-mentioned challenges and suffer from cascading effects. The detection and rectification of such errors remain largely underexplored. To this end, we introduce \textbf{\name{}}, \textbf{G}ated \textbf{R}esidual \textbf{Att}ention for \textbf{V}ideo \textbf{I}nstance \textbf{S}egmentation. Firstly, we leverage a Gumbel-Softmax-based gate to detect possible errors in the current frame. Next, based on the gate activation, we rectify degraded features from its past representation. Such a residual configuration alleviates the need for dedicated memory and provides a continuous stream of relevant instance features. Secondly, we propose a novel inter-instance interaction using gate activation as a mask for self-attention. This masking strategy dynamically restricts the unrepresentative instance queries in the self-attention and preserves vital information for long-term tracking. We refer to this novel combination of Gated Residual Connection and Masked Self-Attention as \textbf{GRAtt} block, which can easily be integrated into the existing propagation-based framework. Further, GRAtt blocks significantly reduce the attention overhead and simplify dynamic temporal modeling. \name{} achieves state-of-the-art performance on YouTube-VIS and the highly challenging OVIS dataset, significantly improving over previous methods. Code is available at \url{https://github.com/Tanveer81/GRAttVIS}.   
\end{abstract}

\section{Introduction}
\label{sec:intro}
Video Instance Segmentation (VIS) \cite{MaskTrackRCNN} is a complex task that requires detecting, segmenting, and tracking all instances or objects within a video sequence. Existing methodologies for VIS can be broadly categorized into offline and online methods. Offline methods \cite{SeqFormer,cheng2021mask2former,MS-STS,VITA,TeViT} process the entire video at once. In contrast, online methods \cite{QueryInst-VIS,STMask,EfficientVIS,VISOLO,InstanceFormer,IDOL,MinVIS,GenVIS} process video sequences frame by frame. The recent emergence of datasets \cite{OVIS,YouTube-VIS-2022} containing lengthy and occluded videos has presented more challenging, real-world scenarios for VIS, driving the development of advanced deployable online models, particularly Detection-Transformer (DETR) \cite{DETR,Deformable-DETR} variants.   
These models primarily rely on frame-level detection and inter-frame association facilitated through tracking \cite{IDOL,MinVIS} or query propagation \cite{EfficientVIS,InstanceFormer,GenVIS,ROVIS}.    

Online models have shown remarkable robustness in processing lengthy and challenging videos and achieved higher accuracy than offline approaches. Their success arises from the capacity to accurately model local complexity, which helps avoid instances from deviating off course over an extended period. Successful modeling of the global context in these cases is achieved by temporal instance association. A DETR-like architecture specifically aids such design endeavors due to its inherent discrete instance representations and their temporal proximity. As a result, instance similarity between frames can be exploited and stitched into a coherent narrative of the instance's evolution throughout the video sequence. Recent methods \cite{MinVIS,IDOL} exploited this characteristic to enhance instance similarity or association between frames. In contrast, several methods \cite{InstanceFormer,GenVIS,ROVIS} continually represent evolving objects in videos by propagating instance queries from frame to frame. 

\begin{figure}
    \centering
    \includegraphics[width=0.99\textwidth]{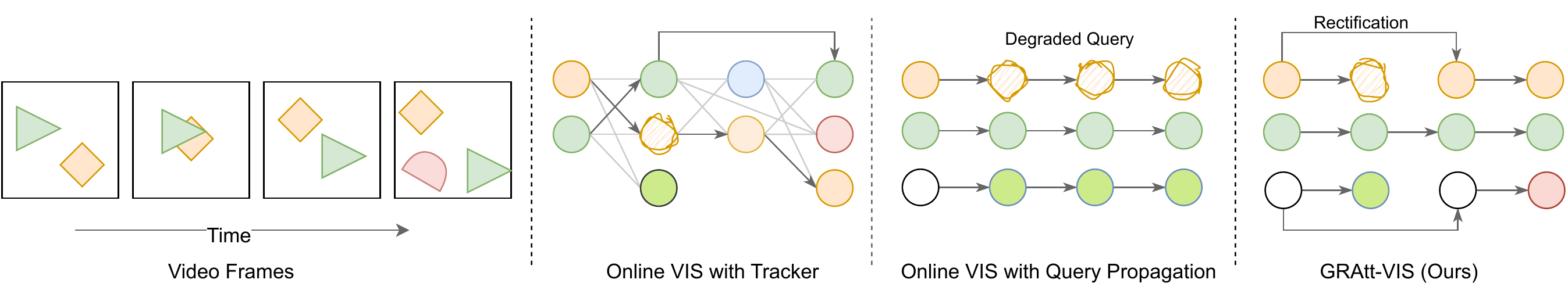}
    \caption{Schematic illustration of trade-offs in previous online VIS. While tracker-based methods suffer from computational complexity, vanilla propagation lacks a decisive decision system. Our \name{} bridges both paradigms with a single network capable of replacing the heuristic association of the tracker with a learned rectified propagation method. }
    \vspace{-0.5em}
    \label{fig:overview}
\end{figure}

Despite the superior performance of online methods compared to offline ones, there remains substantial scope for improvement, as shown in Fig. \ref{fig:overview}. For instance, tracking-based online methods \cite{IDOL,MinVIS,VISOLO,CrossVIS} heavily rely on the heuristic tracker, which significantly deteriorates inference speed and performance with an increased number of objects. Moreover, handcrafted tracking modules limit the network's expressiveness of end-to-end learning and lack scalability across datasets. 
In contrast, query propagation allows the same network to optimize detection and tracking modules without requiring an external tracker. However, tracking in complex videos presents significant challenges, such as short or long-term occlusion, instance appearance or disappearance, and ID switching. We observe that the propagation strategy of \cite{EfficientVIS,InstanceFormer,ROVIS} is helpful when consecutive frames share a considerable similarity. However, in abrupt changes, propagation might accumulate errors like lost tracks, ID switches, or inconsistent masks. Methods like \cite{InstanceFormer,GenVIS} incorporates a memory bank consisting of past representations to mitigate the impact of erroneous propagation.  However, designing an optimal memory queue is a double-edged sword because a small memory size may not facilitate instance recovery, while a large memory may introduce noisy representations. Moreover, integrating a memory queue through cross-attention is resource-intensive and poses optimization challenges due to the \lq irrelevant' and redundant features in the memory bank. 

We posit that effective propagation can be achieved by preserving \emph{relevant} instance queries and rectifying errors from the noisy ones. After interacting with current frame features, instance queries that lose their context or tracks are referred to as \emph{irrelevant queries} going through a \emph{shock}. Shocks include, but are not limited to, situations such as blur, abrupt camera movement, occlusion, and new object appearance. Finding hand-crafted criteria for detecting such irrelevant queries to recover from the shock is hard to formalize exhaustively. As an alternative direction, we aim to incentivize the network to implicitly model the distribution of shock occurrence at the instance level. In the presence of a shock, we argue that previous frame queries are more relevant in maintaining a consistent and accurate instance representation. This bias of earlier frames could \emph{auto-rectify} errors from the current erroneous frame features and makes the queries more robust and stable throughout time. 

\begin{figure}
    \centering
    \includegraphics[width=1\linewidth]{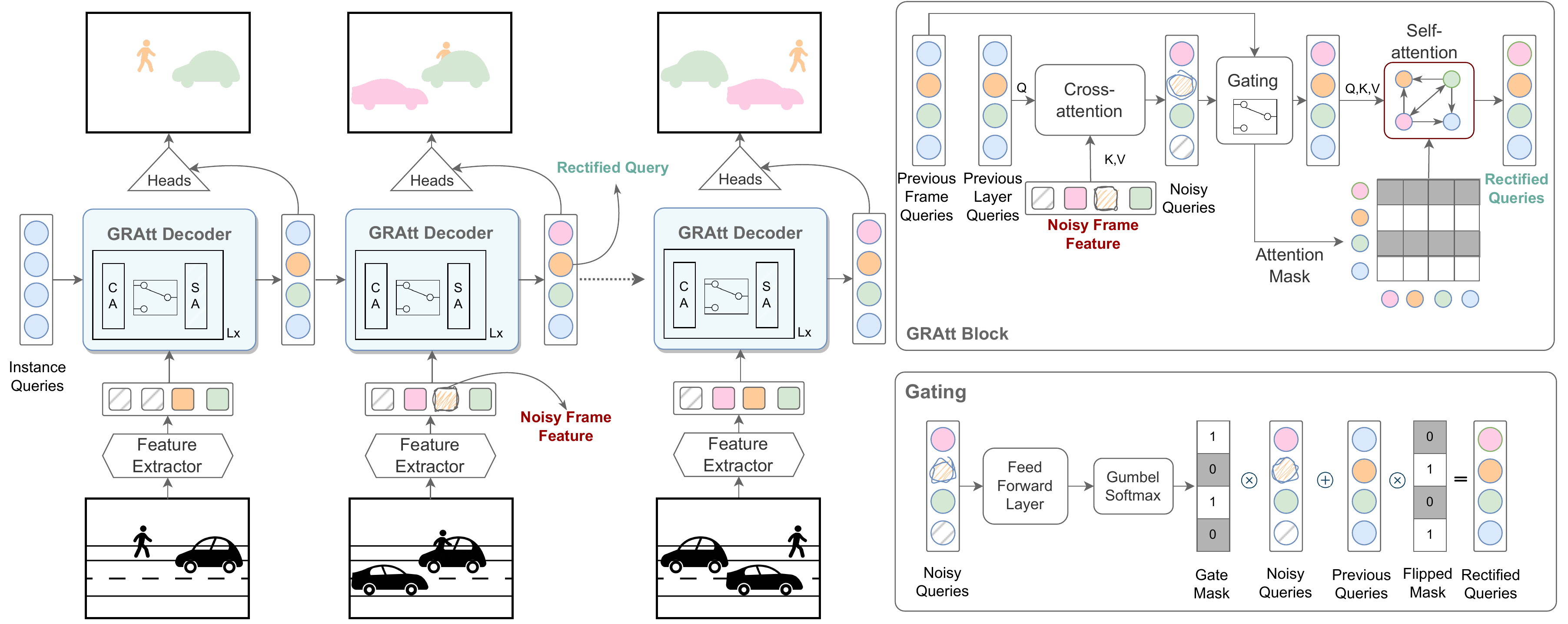}
    \vspace{-1em}
    \caption{\textbf{The architecture of \name{}}. Our GRAtt-decoder is a generic architecture to make the temporal query propagation robust and stable against abrupt noise and shock. This is achieved by a gated residual connection following a masked self-attention. Together they form a Gated Residual Attention (GRAtt) block, which learns to rectify the effect of noisy features and implicitly preserves \lq relevant' instance representations along the temporal dimension. The GRAtt block is a simple replacement for computation-heavy memory and provides superior performance on complex videos across multiple propagation-based VIS frameworks.}
    \vspace{-1em}
    \label{fig:farvis_arch}
\end{figure}

\textbf{Our Contribution:} To this end, we present a \textbf{G}ated \textbf{R}esidual \textbf{Att}ention for \textbf{V}ideo \textbf{I}nstance \textbf{S}egmentation, termed \textbf{\name{}} as shown in Fig. \ref{fig:farvis_arch}. Our approach aims to enhance temporal consistency in query propagation while making it robust against abrupt noise or shock. \textbf{Firstly}, we introduce a Gumbel-Softmax-based \cite{gumble-softmax} residual gating mechanism, which controls query propagation conditioned on the learned relevance of their current representation. Specifically, the residual-gating mechanism learns to decide whether to keep the previous instance query or propagate the current one. This auto-rectifies degraded queries from accumulating noise that could affect current and future predictions. \textbf{Secondly}, we apply gate activation to restrict interaction among the instance queries. To preserve queries perturbed by shock in the current frame, we prevent them from interacting with other instance queries by introducing a novel self-attention mask. At the same time, all relevant queries attend to the global context. This not only helps in query rectification but also sparsifies attention mechanisms and reduces the computational cost. 

Together these two components form the Gated Residual Attention (GRAtt) block in \name{} and improve on three key challenges of VIS: occlusion, new object detection, and robustness against abrupt perturbation in frames. This is achieved via implicitly detecting the shock and rectifying degraded queries. Moreover, \name{} offers enhanced performance while decreasing the computational load, or GFLOPs. Importantly, our method can be integrated ad-hoc into existing propagation-based networks, i.e., InstanceFormer~\cite{InstanceFormer}, GenVIS~\cite{GenVIS}, emphasizing its simplicity and efficiency. \name{} achieves state-of-the-art performance across multiple benchmark VIS datasets, such as, YouTubeVIS-19/21/22 \cite{MaskTrackRCNN,YouTube-VIS-2021,YouTube-VIS-2022} and OVIS \cite{OVIS}. In comparison to the previous best baseline~\cite{GenVIS}, \name{} improves performance in terms of Average Precision (AP) by $1.8\%$ in YTVIS-21, $3.3\%$ on YTVIS-22  long videos, and $0.4\%$ on OVIS. In addition, \name{} reduces the computational effort on average by $37.8\%$ of GFLOPs over the baseline Decoder.

\section{Related Literature}
\label{sec:rel_lit}
Existing VIS methods can be classified into two categories, Offline and Online. Furthermore, we can distinguish online methods into Tracker-based and Query Propagation-based approaches. 

\textbf{Offline-VIS} processes the whole video simultaneously, making future frames available during inference. Earlier offline-VIS incorporated instance mask propagation~\cite{lin2021video,bertasius2020classifying} for temporal connection. Recently, the instance query of the Detection Transformer~\cite{DETR} has been exploited vastly in this paradigm. IFC~\cite{IFC} pioneered this line of research with the very first end-to-end trainable query-based method. VisTR~\cite{VisTR}, SeqFormer~\cite{SeqFormer}, continued this effort by reducing overall complexity and improving performance. They focused on building efficient temporal attention mechanisms to achieve this feat. Mask2Former-VIS~\cite{cheng2021mask2former} further demonstrated that powerful frame-level query-based detectors could compete with contemporary methods with little overhead. TeViT~\cite{TeViT} and VITA~\cite{VITA} went further by limiting the temporal attention with a more effective shifted window mechanism. Finally, EfficientVIS~\cite{EfficientVIS} developed a streamlined heuristic for linking clips, thereby facilitating the processing of longer videos.

 \textbf{Tracker-based Online-VIS} requires heuristic post-processing on top of network detection. MaskTrack R-CNN~\cite{MaskTrackRCNN} associates instances with a track head during inference. Subsequent tracker-based works, CrossVIS~\cite{CrossVIS}, and VISOLO~\cite{VISOLO} improved on top of it by utilizing video-level properties in their training pipeline. Though trained on frame-level, MinVIS~\cite{MinVIS} substantially improves on previous methods by leveraging a powerful object detector Mask2Former~\cite{cheng2021mask2former} and tracking instances with bipartite matching. The current state-of-the-art tracker-based method IDOL~\cite{IDOL}, is built on Deformable-DETR~\cite{Deformable-DETR}. It adopts contrastive learning on the instance queries between frames during training and deploys a heuristic instance matching during inference. 
 
 \textbf{Query Propagation-based Online-VIS} eliminates the need for an external tracker or data association by leveraging the instance queries readily available in Detection Transformer~\cite{DETR} based architectures. TrackFormer~\cite{TrackFormer} initially demonstrated this powerful technique on Multi-Object Tracking and Segmentation~\cite{MOTS} challenge. Later on, this method was adopted by InstanceFormer~\cite{InstanceFormer}, and ROVIS~\cite{ROVIS} for the VIS datasets. InstanceFormer propagated instance queries and reference points of Deformable-DETR~\cite{Deformable-DETR} from frame to frame. On the other hand, ROVIS utilizes the more powerful frame level Mask2Former~\cite{cheng2021mask2former} feature to establish the inter-frame link. Before our work, GenVIS~\cite{GenVIS} held state-of-the-art performance by crafting a decisive video-level training strategy utilizing Mask2Former architecture.

\section{Methodology}
\label{sec:meth}
In this section, we present \name{}, comprising a novel Gated Residual Attention (GRAtt) in the \textbf{Decoder}. We begin by providing a brief background on the query propagation-based VIS methods. Afterward, we present our proposed gating mechanism, its use in residual propagation, and masked self-attention.

\subsection{Query Propagation based VIS}
Recent models, such as GenVIS \cite{GenVIS} and InstanceFormer\cite{InstanceFormer}, use a transformer architecture that processes each frame and passes their instance representation to the next frame. This mechanism is referred to as query propagation. Let video $\mathbf{V} \in \mathbb{R}^{N_f \times H \times W\times3}$ consist of $N_f$ frames of height $H$ and width $W$. At the time $t$, a query propagation-based model extracts feature $f_{t}$ from a frame $x_t \in \mathbf{V}$ through its feature extractor $f_{t} = \Phi(x_{t})$, where $\Phi$ is Mask2Former~\cite{cheng2021mask2former} in GenVIS, and ResNet-50~\cite{ResNet} in InstanceFormer. Afterward, it contextualizes $f_{t}$ through a transformer encoder to be used by a decoder. The decoder uses $N$ instance queries that learn the instance representation. 
 
Both GenVIS and InstanceFormer process each frame independently to obtain current frame features. The instance queries in the decoder attend to current frame features through a cross-attention followed by a self-attention-based contextualization. After processing the current frame, the instance queries are propagated to the subsequent one. The propagation mechanism assigns instances with a particular indexed query throughout the video. Once a query has been associated with an instance, it becomes persistently linked to that query, thereby preventing reassignment to any new objects that may appear later in the video. A new object's appearance is captured by the unallocated queries. During training, Hungarian matching is applied only upon the appearance of a new object. To mitigate noise and re-identify objects in long videos, these models often employ additional memory attention. This feature is implemented by an extra cross-attention module and a memory bank consisting of past instance queries. GenVIS has a transformer on top of Mask2Former, whereas, InstanceFormer relies on Deformable-DETR \cite{Deformable-DETR}. Apart from instance queries, InstanceFormer also propagates reference points and class distribution. \par Despite using the same design principle, GenVIS, and InstanceFormer have different architectural and operational details. We base \name{} primarily on the GenVIS architecture for our experiments as a default setting because of its superior performance over InstanceFormer. Nevertheless, \name{}'s InstanceFormer variant provides a comprehensive assessment of our universality, efficacy, and robustness across architectures. The following section describes our proposed decoder block that replaces previously used memory. In our default settings, the initial feature extractor and the final head remain the same as for GenVIS.

\subsection{GRAtt Decoder}
At the core of our contribution lies a Gumbel-Softmax-based~\cite{gumble-softmax,gumble-softmax_2} gating mechanism within the decoder layer. The decoder takes the instance queries as input and attends to both frame features and instance queries, playing a crucial role in object localization. We aim to assess the relevance of queries to correctly capture instance representation and discard them in case of degradation. This necessitates the deployment of a hard-gating mechanism directly within the decoder since a definitive \lq yes' or \lq no' answer is required to decide whether to propagate the queries through the frames or not.
We leverage the Gumbel-Softmax trick, a technique capable of transforming a continuous distribution into a categorical one. This trick allows us to attain the binary gating output crucial to our method while ensuring end-to-end differentiability. Our gating mechanism learns the distribution of occurrences of abrupt perturbations based on a dataset. Note that we do not provide any explicit supervision to the gating output. Consequently, one can interpret the gate output as an auto-rectifying mechanism targeted toward the most useful instance queries throughout the temporal dynamics. 

\textbf{Gumbel-Softmax based Gating:}
At frame $t$, we consider $q_{t}^{i}\in \mathbb{R}^{C}$ to be the $i^{th}$ object query, where $i \in \{1,..,N\}$. We place the gating after each cross-attention layer so that a query can accumulate necessary frame information. Consequently, we want to assess the relevancy of the query feature with respect to the current frame. At decoder layer $l  \in \{1,..,L\}$, after the cross-attention, for $q_{t}^{i}$ we obtained its corresponding gate signal $g_t^{i,l}$ as
\begin{equation}
    g_t^{i,l} = W_{\mbox{gate}}\left(q_t^{i,l}\right),\   \   g_t^{i,l} \in \mathbb{R}\  \mbox{and}\  W_{\mbox{gate}} \in \mathbb{R}^{1\times C} 
\end{equation}
 where $W_{\mbox{gate}}$ is a linear projection layer with output dimension $1$. 

We want to obtain, a categorical variable $G_t^{i,l}$ with probabilities ${\pi_1}_t^{i,l} = \sigma(g_t^{i,l})$ and ${\pi_0}_t^{i,l} = 1- \sigma(g_t^{i,l})$, where $\sigma$ is the sigmoid operation. We can reparameterize the sampling process of $G_t^{i,l}$ using the Gumbel-Max trick as follows:
\begin{equation}
G_t^{i,l}={\arg \max}_k \left\{\log \left({\pi_k}_t^{i,l}\right)+g_k:k=0,1\right\}
\end{equation}

Here, $\{g_k\}_{k={0,1}}$ are i.i.d. random variables sampled from $Gumbel(0,1)$. Due to the discontinuous nature of the argmax operation, we approximate $G_t^{i,l}$ with a differentiable softmax function. The differentiable sample $\hat{G}_t^{i,l}$ obtained from the Gumbel-Softmax relaxation is given by:
\begin{equation}
\hat{G}_t^{i,l}=\frac{\exp \left(\left(\log \left({\pi_1}t^{i,l}\right)+g_1\right)/{\tau}\right)}{\Sigma_{k \in{0,1}} \exp \left(\left(\log \left({\pi_k}_t^{i,l}\right)+g_k\right)/\tau\right)}
\label{eqn:gumbel-softmax}
\end{equation}

Following the recommendations in the literature~\cite{gumble-softmax_2}, we set the softmax temperature $\tau$ to $2 / 3$. 

\textbf{Gated Residual Connection:}
By convention, instance queries flagged as `zero' in the gating mechanism are considered `irrelevant' and do not propagate through the current frame. If propagated, the `irrelevant' features could accumulate erroneous contextualization of instances resulting in lost tracks. Instead, a residual connection from the previous frame supplies a meaningful representation of instance queries. Our novel gated residual connection is placed between the decoder's cross-attention and self-attention layers. This connection can be described as 
\begin{equation}
q_{t}^{i,l+1} = \left\{\begin{array}{ll}
q_{t}^{i,l} & \text { if } \hat{G}_t^{i,l}=1 \\
q_{t-1}^{i,L} & \text { if } \hat{G}_t^{i,l}=0
\end{array} \right.
\label{eq:rescon}
\end{equation}

Eq. \ref{eq:rescon} allows propagation of all `relevant' queries to the subsequent layers, whereas degraded queries are rectified with their preceding temporal counterpart. Such residual connection effectively manages the flow of queries in a video, including instance reappearance in case of occlusion and the preservation of robust object representation. Although the residual connection is placed between layers, the correction terms are retrieved from the past frame instead of the past layer. This configuration establishes effective frame-to-frame continuation and provides additional expressiveness during intra-frame processing. Consequently, degraded features are rectified through the residual connection with the \lq relevant' ones prior to occlusion or distortion.
Furthermore, unallocated instance queries might be available during processing and can easily bind to new objects. Moreover, unallocated instance queries reduce duplicate proposals and act as an implicit \textit{Non-Maximal Suppression}. Thus, residual gating mitigates two major challenges in tracking without an explicit tracker or a memory module. The ablation on different design choices of the place of the residual connection among inter-frame, inter-layer, and inter-attention is visualized in Fig. \ref{fig:rescon} and justified in ablation Tab. \ref{tab:my_label1} in the supplementary.

\textbf{Masked Self-attention:}
Upon computation of $\hat{G}_t^{i,l}$ for each query, we introduce a novel \textit{masked self-attention} by computing a mask for the subsequent self-attention layer. This mask allows object queries with a corresponding gate value of `one' to engage in global self-attention with the rest of the queries. In contrast, object queries yielding a `zero' gate output are masked out, effectively removing them from self-attention considerations.

A gate value of `zero' indicates a lack of useful frame features for a particular instance query. We argue that obtaining additional information from self-attention is unlikely in such scenarios since the relevant instance queries were propagated through the current frame features. Further, there is the possibility of noisy frame feature injection through self-attention. Therefore, by masking out such queries, we adopt a greedy strategy to prefer undistorted instance representations over the most updated ones. Conversely, global attention to \lq relevant' queries with gate value \lq one' enables them to contextualize useful information from other queries.

Simultaneously, this approach sparsifies the number of queries, thereby leading to a reduction in computation time. In contrast to existing methods, ~\cite{IDOL, GenVIS} whose computation increases with an increasing number of objects, \name{} counter-intuitively requires less time because of increased chances of occlusion and thereby reduced attention computation.

Let's denote $\boldsymbol{f}^l_q=[q^{1,l}_{t}, q^{2,l}_{t},\cdots q^{N,l}_{t}]\in \mathbb{R}^{N \times C}$ to be the set of all queries at the $t^{\mbox{th}}$ frame, \textit{masked self-attention} is computed as
\begin{equation}
    \boldsymbol{f}^{l+1}_q = \operatorname{softmax}(\boldsymbol{Q}_{l}\boldsymbol{K}^{T}_{l}+ \boldsymbol{M}_{l})\boldsymbol{V}_{l} + \boldsymbol{f}^l_q
\end{equation}
$\boldsymbol{Q}_{l}, \boldsymbol{K}_{l}, \boldsymbol{V}_{l} \in \mathbb{R}^{N \times C}$ refers to $N \times C$-dimensional query, key and value features at the $l^{\mbox{th}}$ layer under linear transformation $f_Q(\cdot)$, $f_K(\cdot)$ and $f_V(\cdot)$ of $\boldsymbol{f}^l_q$, respectively. The attention mask $\boldsymbol{M}_{l}$ between $i^{\mbox{th}}$ and $j^{\mbox{th}}$ instance query is computed as
\begin{equation}
   \boldsymbol{M}_{l}(i, j)=\left\{\begin{array}{ll}
0 & \text { if } \hat{G}_t^{i,l}=1\\
-\infty & \text { if } \hat{G}_t^{i,l}=0
\end{array} \right. 
\end{equation}

Together Gated Residual Connection and Masked Self-attention constitute a Gated Residual Attention block, which maintains the integrity of object representation across frames and layers, thereby enhancing the robustness and performance of our model. At the same time, the GRAtt decoder reduces the computational cost thanks to attention sparsification and no additional memory. We justify our design choice in comparison with other attention mask configurations (c.f. Fig. \ref{fig:attn}) in Tab. \ref{tab:my_label2} in the supplementary document. The final layer queries at each time frame go through a head identical to GenVIS to generate the final segmentation. 

\section{Experiments}
\label{sec:exp}
We present \name{} as a robust module targeted to improve the recently developed propagation-based frameworks. Our proposed design is incorporated on top of two such architectures, namely GenVIS and InstanceFormer, and tested on the most prevalent datasets.

\textbf{Datasets:}
We experimented on four benchmark datasets, YouTubeVIS(YTVIS)-19/21/22 \cite{YouTube-VIS-2021, YouTube-VIS-2022} and OVIS \cite{OVIS}. YTVIS is an evolving dataset with three iterations from 2019, 2021, and 2022. It focuses on segmenting and tracking video objects with 40 predefined categories. The dataset's complexity has increased with the introduction of more intricate, longer videos containing complex object trajectories. Despite the YTVIS 2021 and 2022 versions sharing an identical training set, additional 71 long videos have been introduced to the validation set of the 2022 version. In comparison, OVIS is a more recent dataset comprising 25 specified categories, presenting a more challenging setup with significantly higher occlusion. 

\textbf{Implementation Details:}
We use identical hyperparameters and data augmentations specified in GenVIS and InstanceFormer while incorporating \name{}. Since our objective is to assess the contribution of the GRAtt decoder, we did not use a memory module as proposed in GenVIS or InstanceFormer. The distributed training was conducted with four Nvidia-RTX-A6000 GPUs. Please note we did not employ post-processing or inter-frame matching during the inference stage. To facilitate the reproducibility of the results, we include the code in the supplementary. Because InstanceFormer has the self-attention before the cross-attention, we augment a conditional convolution layer in the beginning to generate gate activation. We trained two versions of both InstanceFormer and GenVIS, with and without memory, using their publicly available repositories for our ablation.
\begin{table*}
\centering
\resizebox{0.95\linewidth}{!}
{ 
\begin{tabular}{@{}c|lc|ccccc|ccccc@{}}
\toprule
\multicolumn{3}{c|}{\multirow{2}{*}{Method}}                        &   \multicolumn{5}{c|}{YouTube-VIS 2019} & \multicolumn{5}{c}{YouTube-VIS 2021}\\
\multicolumn{3}{l|}{}                                               &          AP        & AP$_{50}$ & AP$_{75}$ & AR$_1$    & AR$_{10}$ & AP        & AP$_{50}$ & AP$_{75}$ & AR$_1$    & AR$_{10}$ \\
    \midrule
    \midrule
    \multirow{6}{*}{\rotatebox{90}{Offline}}
    & \multicolumn{2}{l|}{EfficientVIS~\cite{EfficientVIS}}         &       37.9      & 59.7      & 43.0      & 40.3      & 46.6& 34.0      & 57.5      & 37.3      & 33.8      & 42.5  \\
    & \multicolumn{2}{l|}{IFC~\cite{IFC}}                           &  41.2      & 65.1      & 44.6      & 42.3      & 49.6                                                                   & 35.2      & 55.9      & 37.7      & 32.6      & 42.9  \\
    & \multicolumn{2}{l|}{Mask2Former-VIS~\cite{cheng2021mask2former}}   &  46.4      & 68.0      & 50.0      & -         & -  & 40.6      & 60.9      & 41.8      & -         & -     \\
    & \multicolumn{2}{l|}{TeViT~\cite{TeViT}}                       &  46.6      & 71.3      & 51.6      & 44.9      & 54.3   & 37.9      & 61.2      & 42.1      & 35.1      & 44.6  \\
    & \multicolumn{2}{l|}{SeqFormer~\cite{SeqFormer}}               &  47.4      & 69.8      & 51.8      & 45.5      & 54.8                                                                   & 40.5      & 62.4      & 43.7      & 36.1      & 48.1  \\
    & \multicolumn{2}{l|}{VITA~\cite{VITA}}                         &  \textbf{49.8}      & \textbf{72.6}      & \textbf{54.5}      & \textbf{49.4}      & \textbf{61.0}& \textbf{45.7}      & \textbf{67.4}      & \textbf{49.5}      & \textbf{40.9}      & \textbf{53.6}  \\
    \midrule
    \multirow{7}{*}{\rotatebox{90}{Online}}
    & \multicolumn{2}{l|}{CrossVIS~\cite{CrossVIS}}                 &  36.3      & 56.8      & 38.9      & 35.6      & 40.7& 34.2      & 54.4      & 37.9      & 30.4      & 38.2  \\
    & \multicolumn{2}{l|}{VISOLO~\cite{VISOLO}}                     &  38.6      & 56.3      & 43.7      & 35.7      & 42.5 & 36.9      & 54.7      & 40.2      & 30.6      & 40.9  \\
    & \multicolumn{2}{l|}{ROVIS~\cite{ROVIS}}                     &  45.5& 63.9& 50.2 &41.8& 49.5 & - & - & - & - & - \\
    & \multicolumn{2}{l|}{InstanceFormer~\cite{InstanceFormer}}                     &  45.6 & 68.6 & 49.6 & 42.1 & 53.5 & 40.8 & 62.4 & 43.7 & 36.1 & 48.1 \\
    & \multicolumn{2}{l|}{MinVIS~\cite{MinVIS}}                     & 47.4      & 69.0      & 52.1      & 45.7      & 55.7 & 44.2      & 66.0      & 48.1      & 39.2      & 51.7  \\
    & \multicolumn{2}{l|}{IDOL~\cite{IDOL}}                         &  49.5      & \textbf{74.0}      & 52.9      & 47.7      & 58.7& 43.9      & 68.0     & 49.6      & 38.0      & 50.9  \\
    & \multicolumn{2}{l|}{GenVIS~\cite{GenVIS}}       &  50.0 & 71.5 & 54.6 & \textbf{49.5} & \textbf{59.7}& 47.1 & 67.5 & {51.5} & {41.6} & {54.7} \\
    
    & \multicolumn{2}{l|}{\textbf{\name{} (Ours)}}  & \textbf{50.4} & 70.7 & \textbf{55.2} & 48.4 & 58.7& \textbf{48.9} & \textbf{69.2} & \textbf{53.1} & \textbf{41.8} & \textbf{56.0} \\
    \bottomrule
    \end{tabular}
} 
\vspace{-2mm}
\caption{Quantitative performance of \name{} compared to previous methods on YTVIS-19/21 datasets. \name{} achieves state-of-the-art performance on both datasets with ResNet-50 backbone.}
\vspace{-3mm}
\label{tab:ytvis2019_2021}
\end{table*}
\begin{table*}
\centering
\resizebox{0.95\linewidth}{!}
{ 
\begin{tabular}{@{}c|lc|ccccc|ccccc@{}}
\toprule
\multicolumn{3}{c|}{\multirow{2}{*}{Method}} &  \multicolumn{5}{c|}{OVIS} & \multicolumn{5}{c}{YouTube-VIS 2022}\\
\multicolumn{3}{l|}{}  &     AP       & AP$_{50}$ & AP$_{75}$ & AR$_1$    & AR$_{10}$ & AP        & AP$_{50}$ & AP$_{75}$ & AR$_1$    & AR$_{10}$ \\
    \midrule
    \midrule
    \multirow{3}{*}{\rotatebox{90}{Offline}}
    & \multicolumn{2}{l|}{SeqFormer~\cite{SeqFormer}}    & 15.1 &31.9& 13.8 &10.4 &27.1     & -&-&-&-&- \\
    & \multicolumn{2}{l|}{TeViT~\cite{TeViT}}      & 17.4      & 34.9          & 15.0          & 11.2          & 21.8   & -&-&-&-&-  \\
    & \multicolumn{2}{l|}{VITA~\cite{VITA}}    & \textbf{19.6}      & \textbf{41.2}          & \textbf{17.4}          & \textbf{11.7}          & \textbf{26.0}& \textbf{32.6}      & \textbf{53.9}      & \textbf{39.3}      & \textbf{30.3}      & \textbf{42.6} \\
    
    \midrule
    \multirow{9}{*}{\rotatebox{90}{Online}}
    & \multicolumn{2}{l|}{CrossVIS~\cite{CrossVIS}}  & 14.9 & 32.7  & 12.1  & 10.3  & 19.8 &  
    -&-&-&-&-  \\
    & \multicolumn{2}{l|}{VISOLO~\cite{VISOLO}}     & 15.3      & 31.0          & 13.8          & 11.1          & 21.7 & 
    -&-&-&-&-  \\
    & \multicolumn{2}{l|}{InstanceFormer~\cite{InstanceFormer}}     & 20.0 & 40.7 & 18.1 & 12.0 & 27.1 & 
    32.0      & 55      & 34.5      & 29.5     & 38.3  \\
    & \multicolumn{2}{l|}{ROVIS~\cite{ROVIS}}     & 30.2 &53.9& 30.1 &13.6 &36.3 & 
    -    & -    & -     & -     & -  \\
    & \multicolumn{2}{l|}{MinVIS~\cite{MinVIS}}     & 25.0      & 45.5          & 24.0          & 13.9          & 29.7 & 
    33.1      & 54.8      & 33.7      & 29.5     & 36.6  \\
    & \multicolumn{2}{l|}{IDOL~\cite{IDOL}}   & 30.2      & 51.3          & 30.0          & 15.0          & 37.5 & 
    -&-&-&-&-  \\
    & \multicolumn{2}{l|}{GenVIS~\cite{GenVIS}} & 35.8 & \textbf{60.8}   & 36.2    & 16.3    & 39.6 & 
    37.5      & \textbf{61.6}      & 41.5      & 32.6      & 42.2 \\
    & \multicolumn{2}{l|}{\textbf{\name{} (Ours)}}  & \textbf{36.2} & \textbf{60.8} & \textbf{36.8} & \textbf{16.8} & \textbf{40.0} & \textbf{40.8}      & 60.1      & \textbf{45.9}      & \textbf{35.7}      & \textbf{46.9} \\
    \bottomrule
    \end{tabular}
} 
\vspace{-2mm}
\caption{Performance comparison of \name{} with recently developed VIS frameworks on the most challenging Occluded (OVIS) and Long (YTVIS-22) Video Instance Segmentation data-sets. The evaluation of SeqFormer is taken from IDOL.
}
\vspace{-3mm}
\label{tab:ytvis2022_OVIS}
\end{table*}

\subsection{Main Results}
 \name{} consistently demonstrates superior performance across all variations of the YTVIS and OVIS datasets. Notably, \name{}, sets a new state-of-the-art AP score across datasets.

\textbf{YTVIS 19 \& 21:}
Table \ref{tab:ytvis2019_2021}, compares the performance of \name{} with previous methods on both YTVIS-19 and 21 datasets.
We observe comparable performance for YTVIS-19 and attribute it to label inaccuracy which was improved in the 2021 version. 
Moreover, YTVIS-21 added more challenging videos on top of YTVIS-19, which contains mostly short videos with gradual changes. \name{} outperforms previous state-of-the-art GenVIS by $1.8\%$ AP, setting a new benchmark on relatively more challenging YTVIS-21.           

\textbf{YTVIS 22:}
Table \ref{tab:ytvis2022_OVIS}, illustrates the performance of \name{} on the YTVIS-22 dataset. Again, \name{} consistently performs better than all earlier models. \name{} elevates the overall AP by $3.3\%$ on top of GenVIS, thereby setting a new benchmark. Note that YTVIS-22 is the most challenging dataset in YTVIS, containing longer and more complex videos. This performance improvement, coupled with reduced computational load, e.g., GFLOPs reduction by $37.8\%$ in the VIS decoder, underscores the effectiveness of our propagation mechanism. The enhanced performance on longer videos also validates the efficacy of the proposed residual propagation and masked self-attention strategy in preserving the relevant object features throughout long video sequences. 

\textbf{OVIS:}
Lastly, we evaluate our method on the highly challenging OVIS dataset in Table \ref{tab:ytvis2022_OVIS}. We observe a similar trend in \name{}'s superior performance as in YTVIS-22. We achieve a $0.4\%$ improvement over GenVIS, which has a memory bank. When comparing to GenVIS without this memory bank, we observe $1.2\%$ gain( Table. \ref{tab:mem_gate_ablation}), suggesting our proposed propagation mechanism is a better and more lightweight alternative to the conventional memory-based methods.

\begin{figure*}[t!]
\centering
\begin{tabular}{cccc}
\includegraphics[width=.22\linewidth]{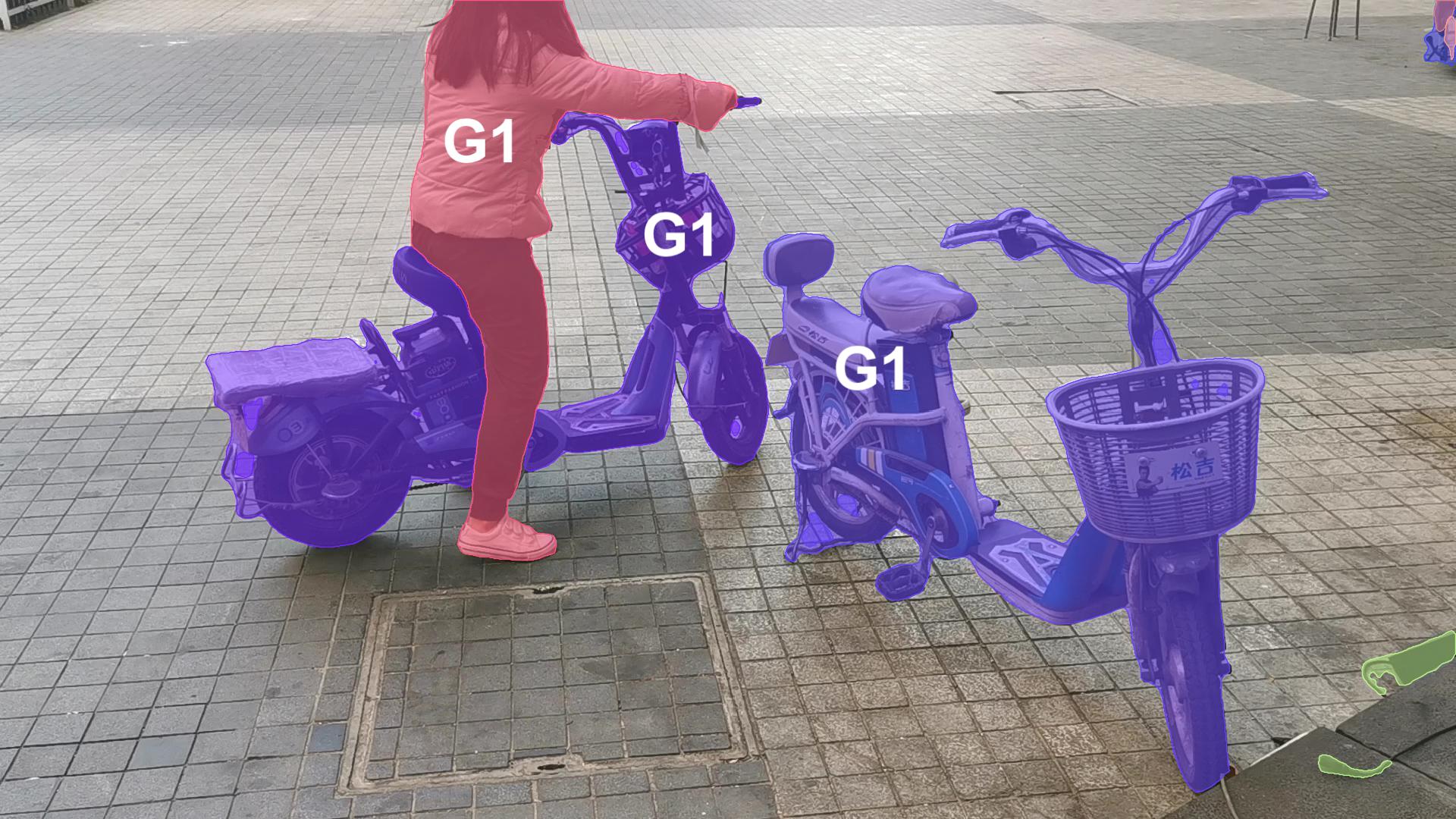} &  
\includegraphics[width=.22\linewidth]{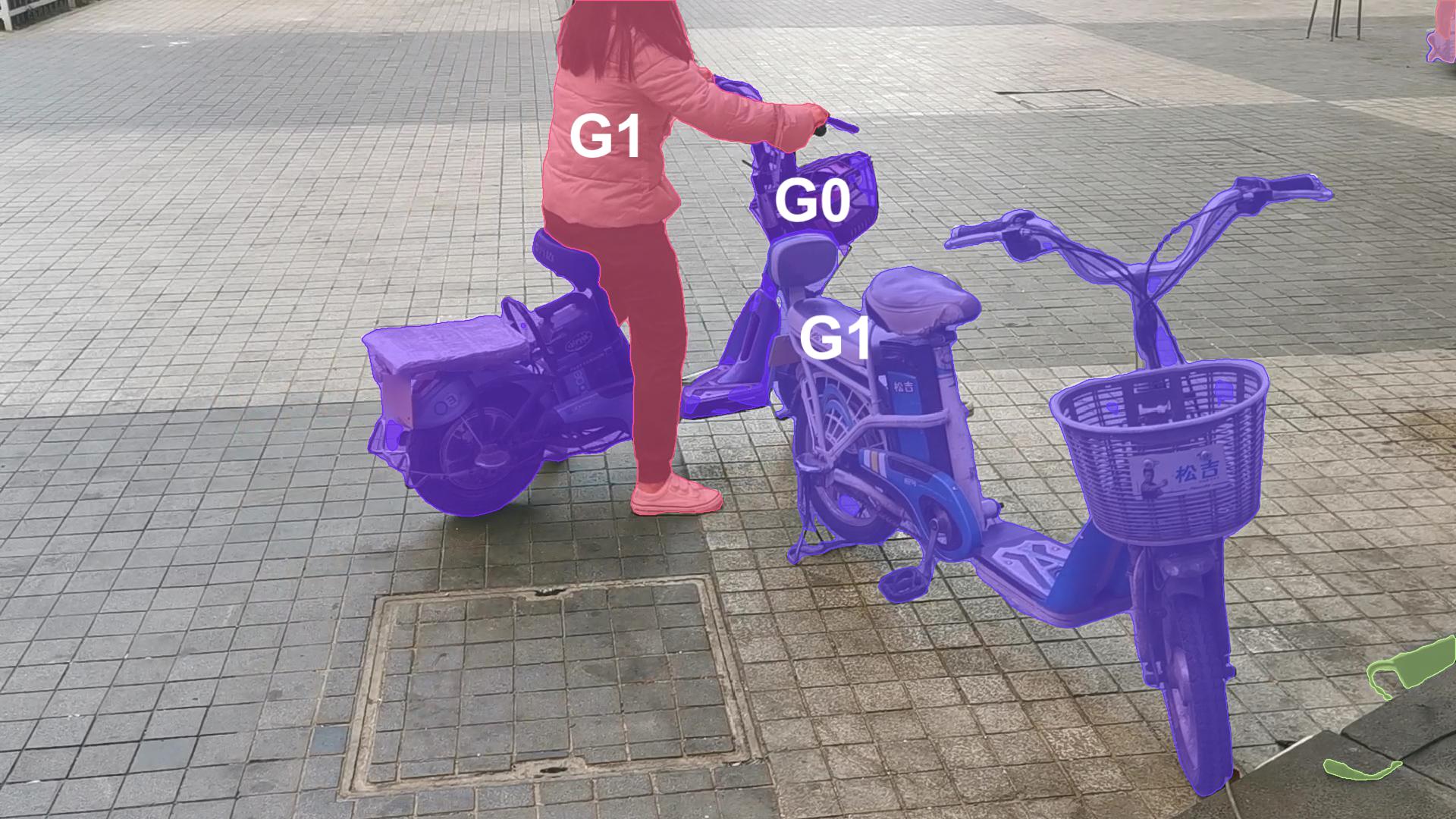} & 
\includegraphics[width=.22\linewidth]{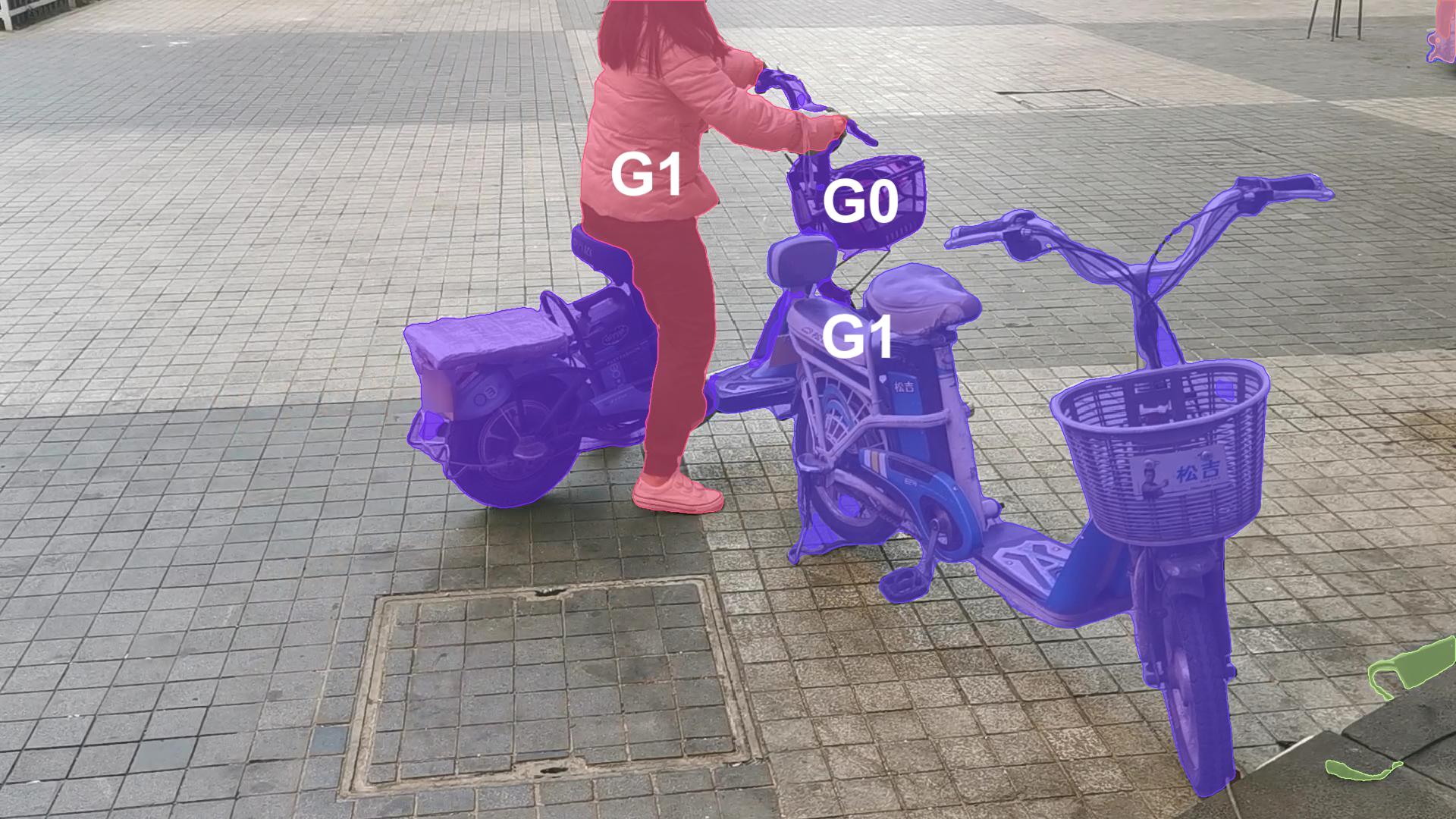} & 
\includegraphics[width=.22\linewidth]{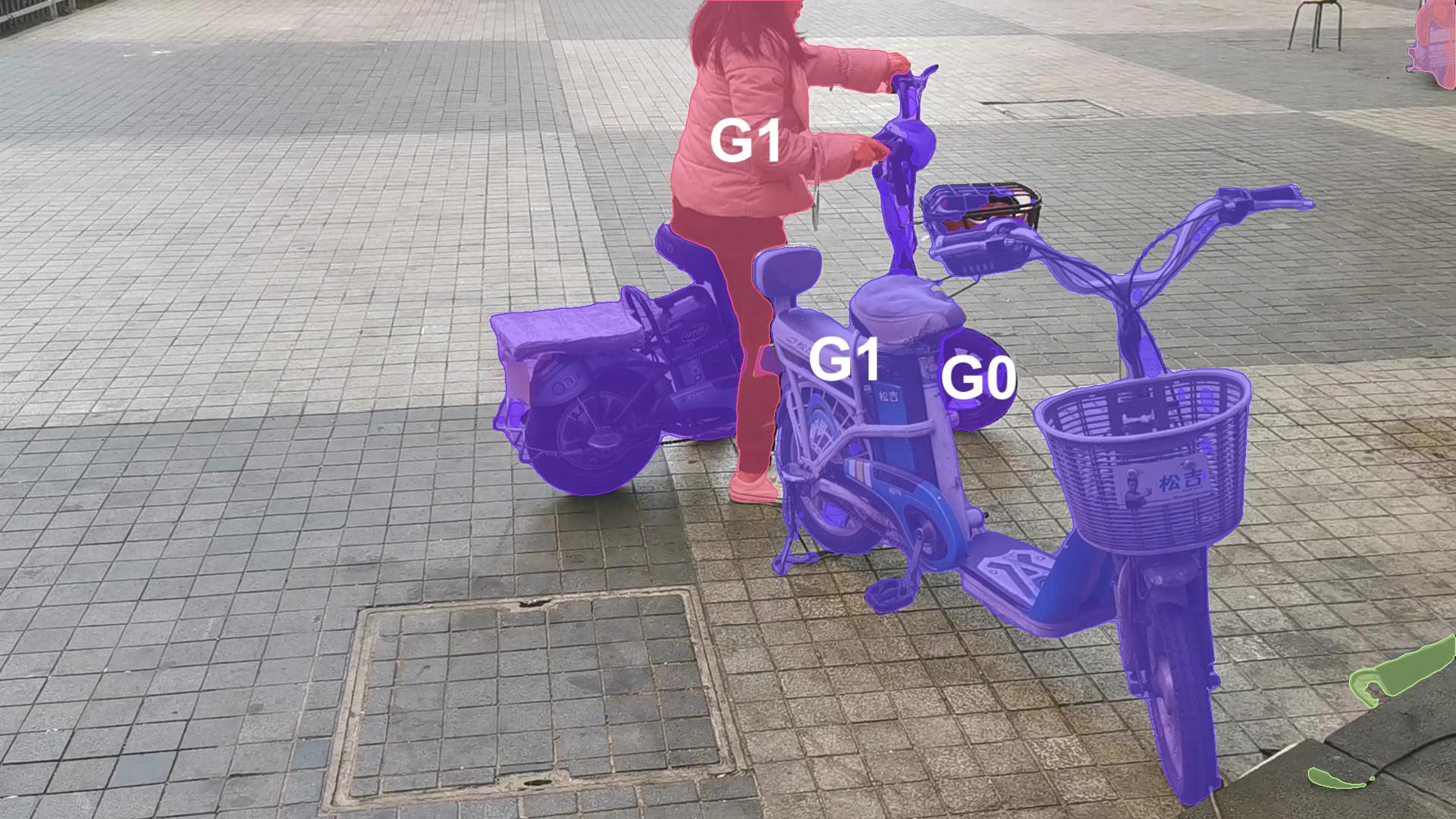} \\
\includegraphics[width=.22\linewidth]{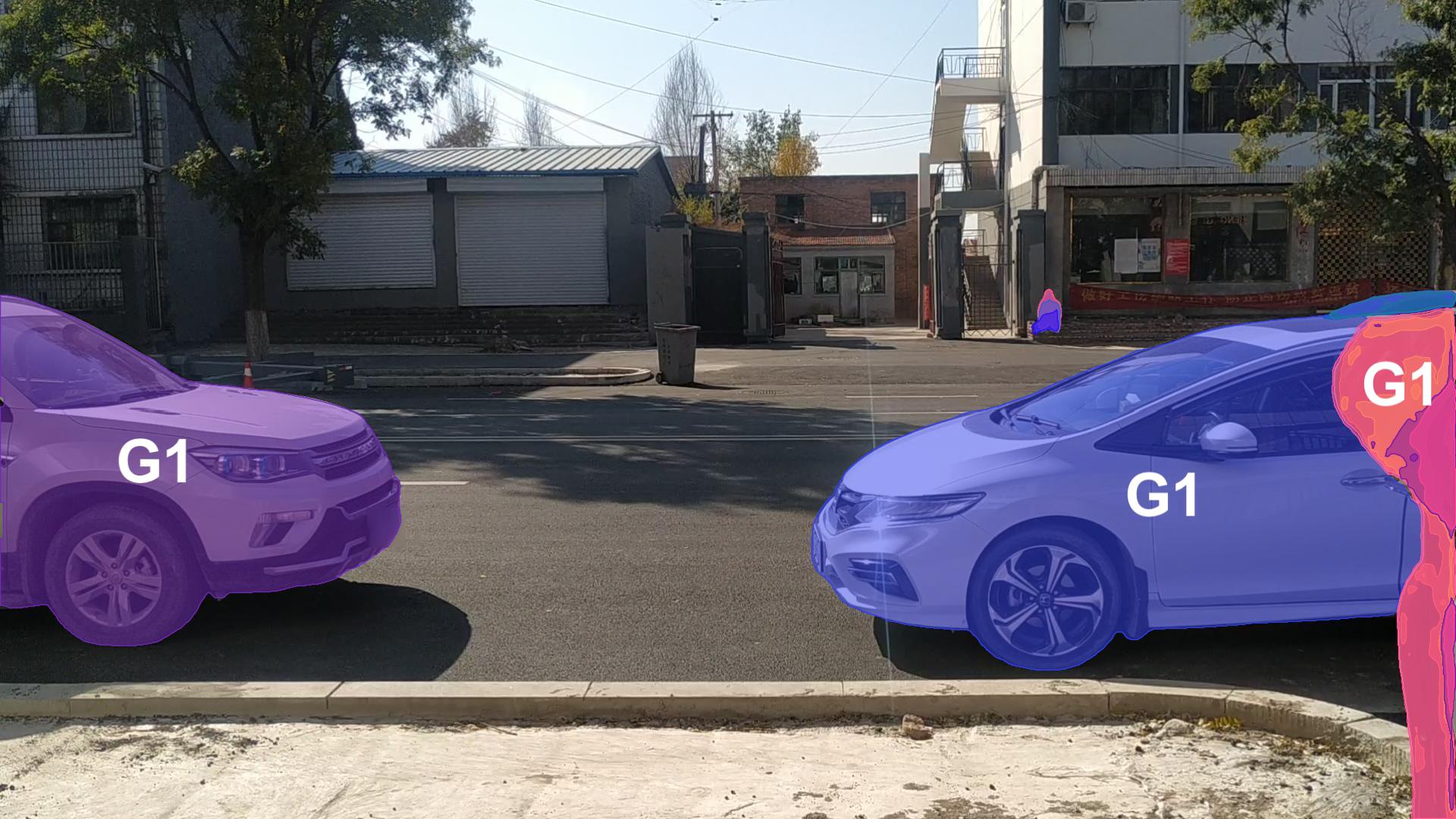} &  
\includegraphics[width=.22\linewidth]{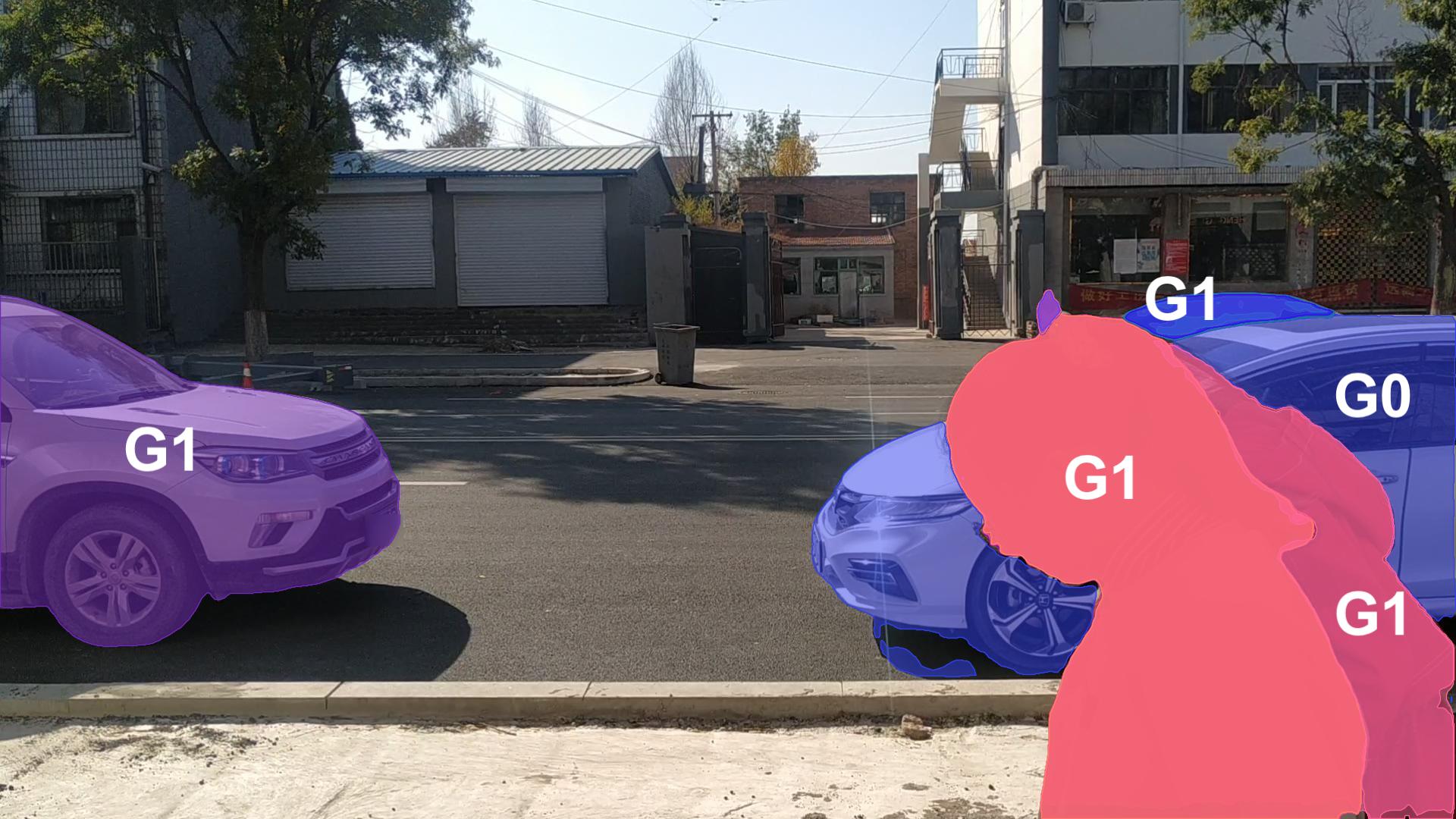} & 
\includegraphics[width=.22\linewidth]{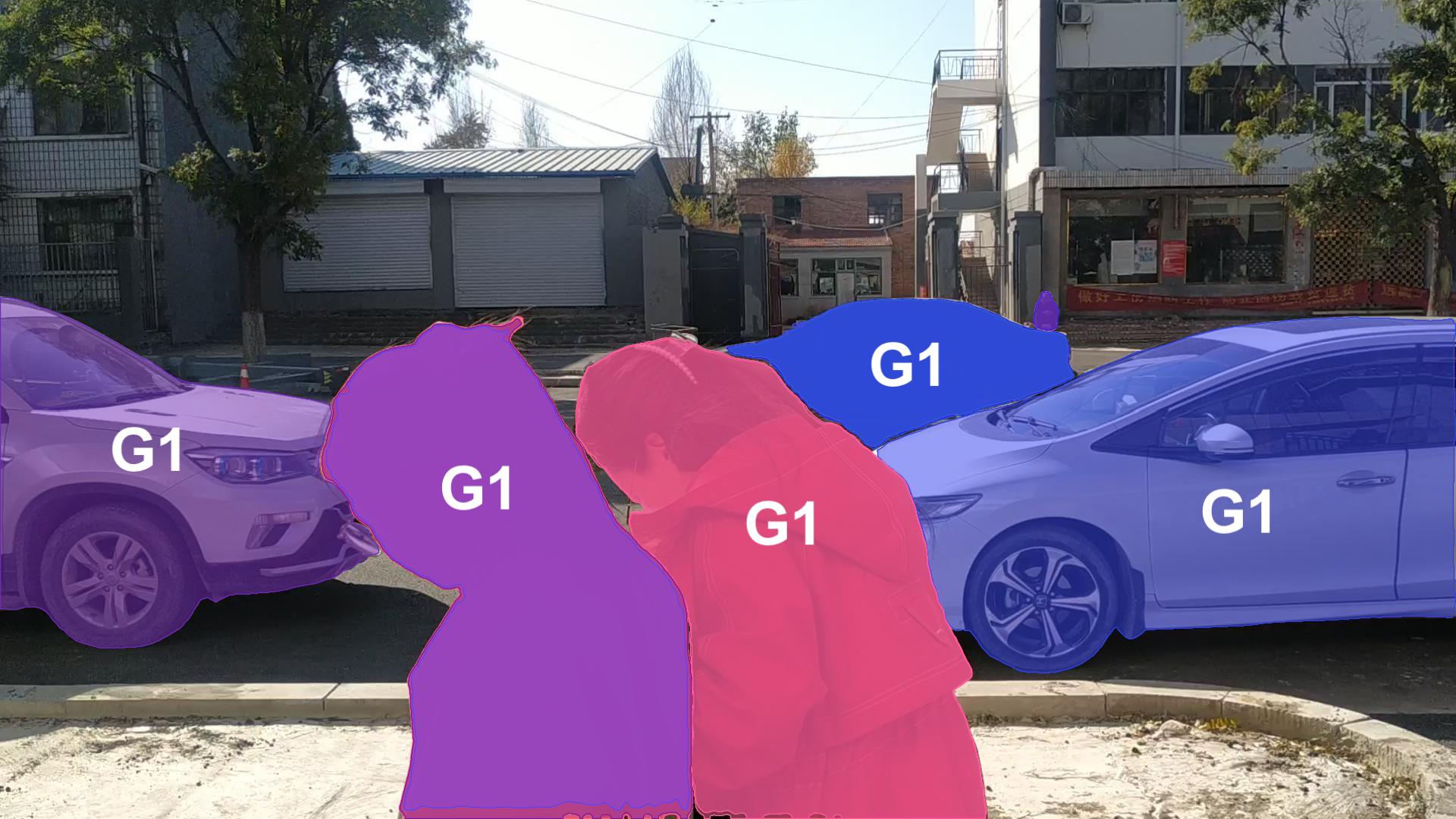} & 
\includegraphics[width=.22\linewidth]{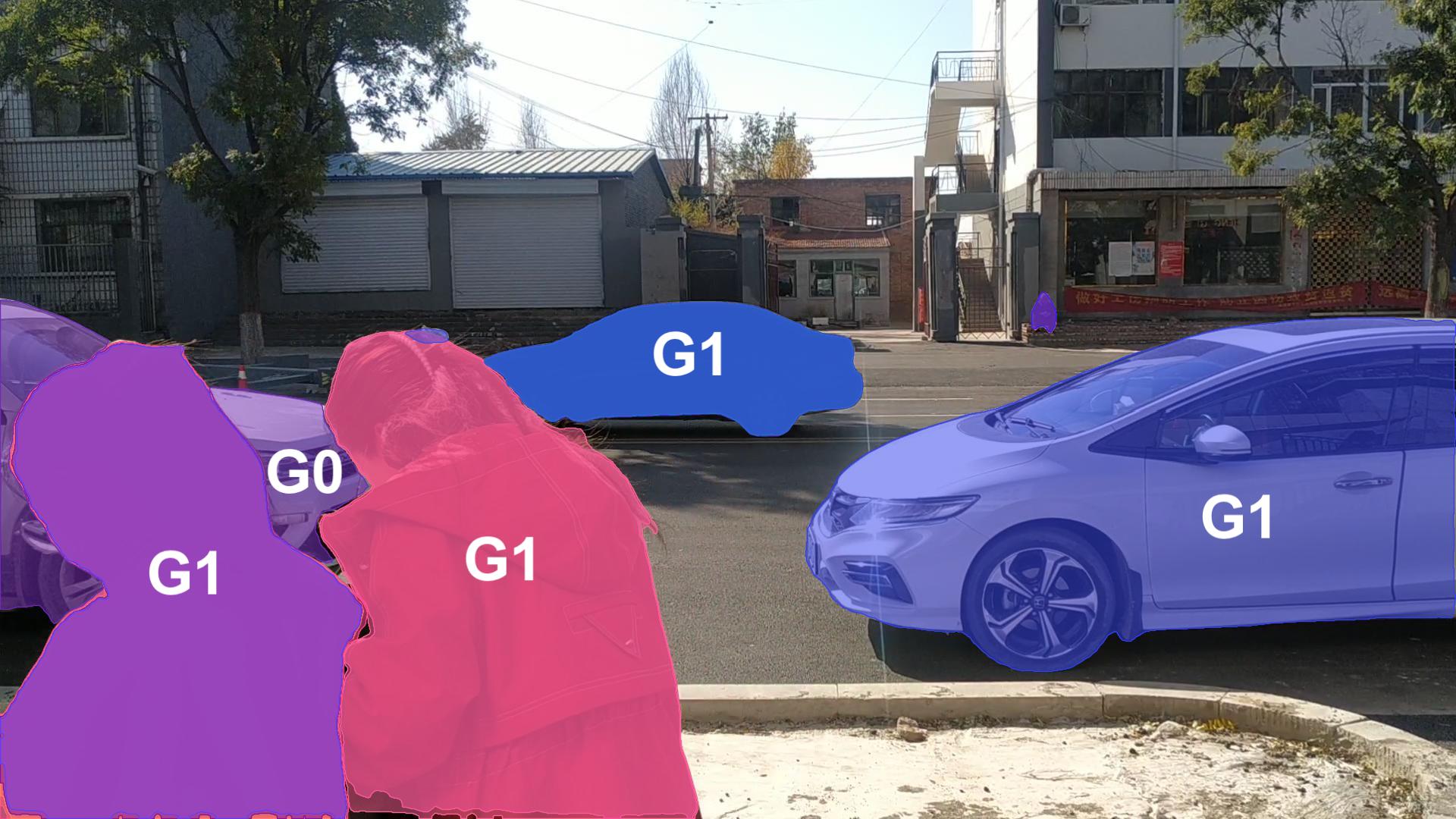}  \\
\includegraphics[width=.22\linewidth]{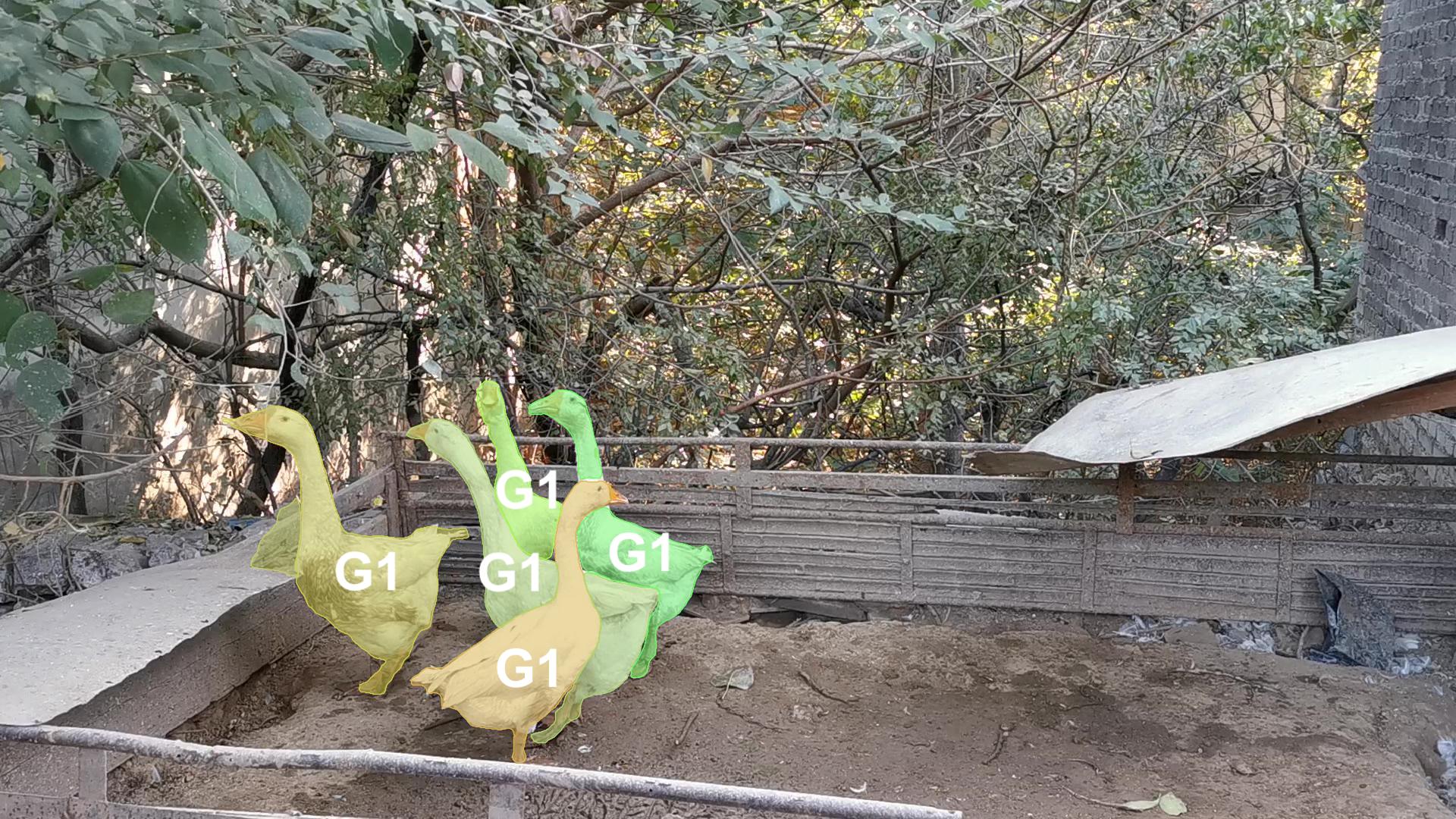} &  
\includegraphics[width=.22\linewidth]{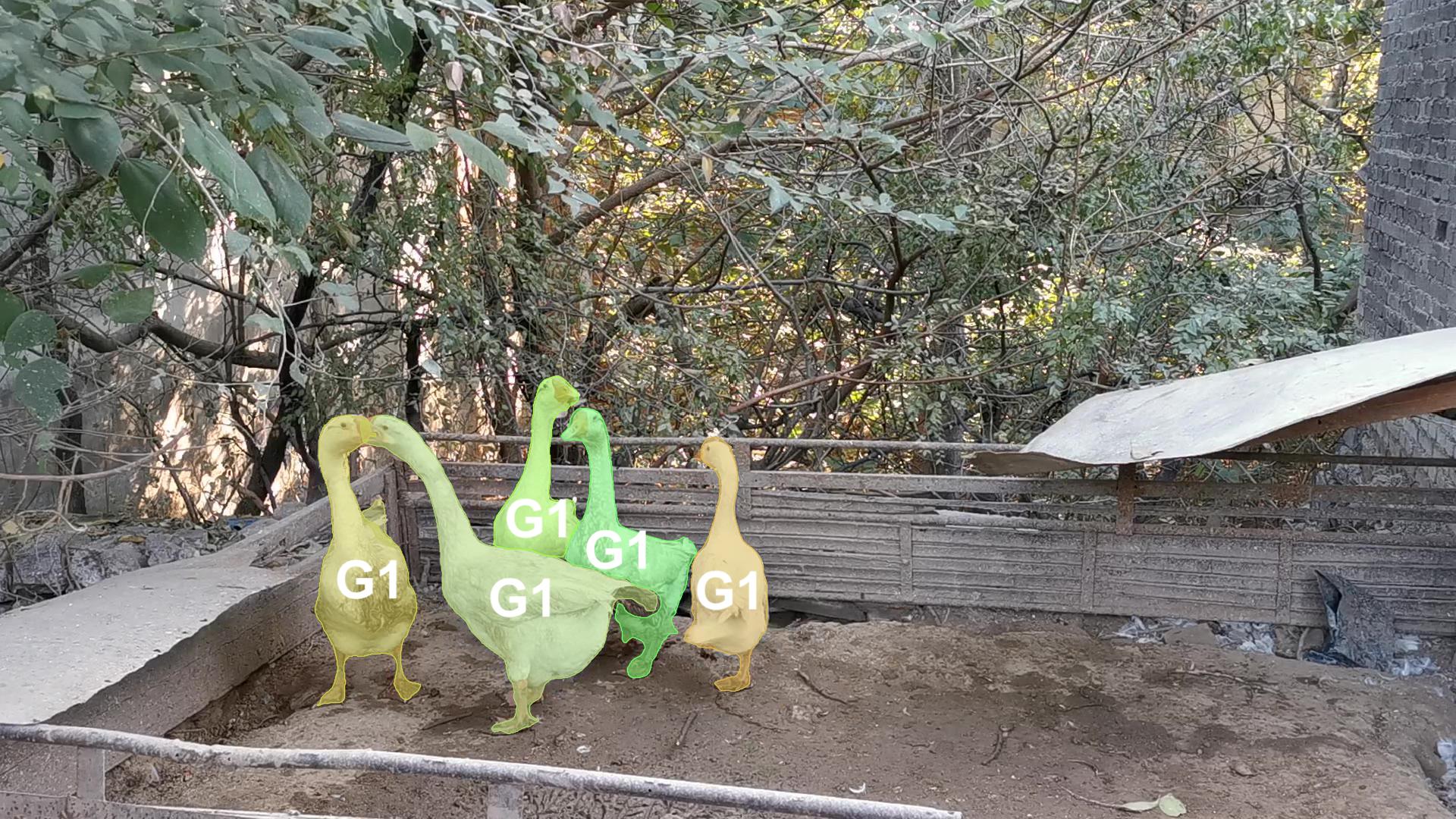} & 
\includegraphics[width=.22\linewidth]{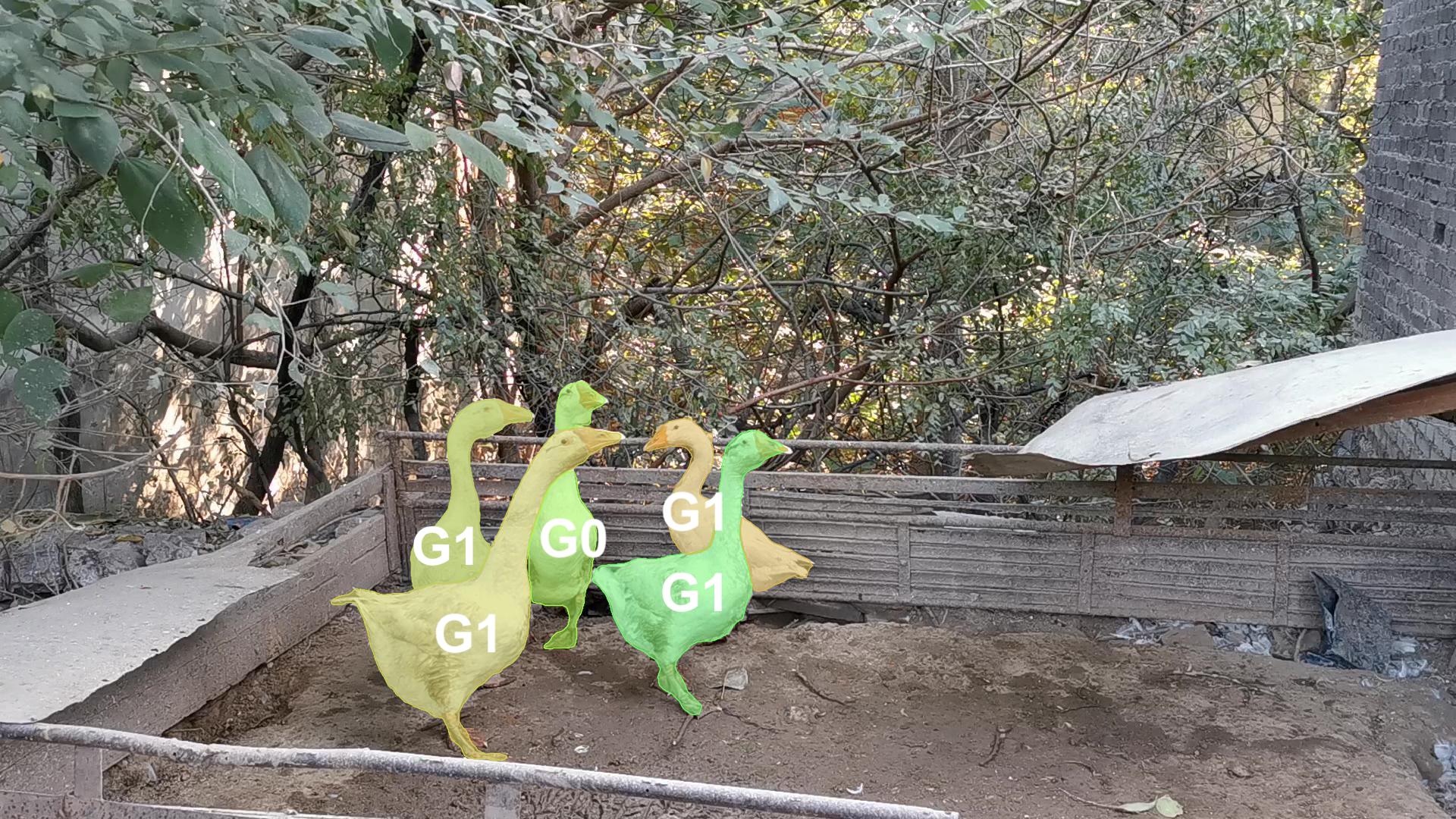} & 
\includegraphics[width=.22\linewidth]{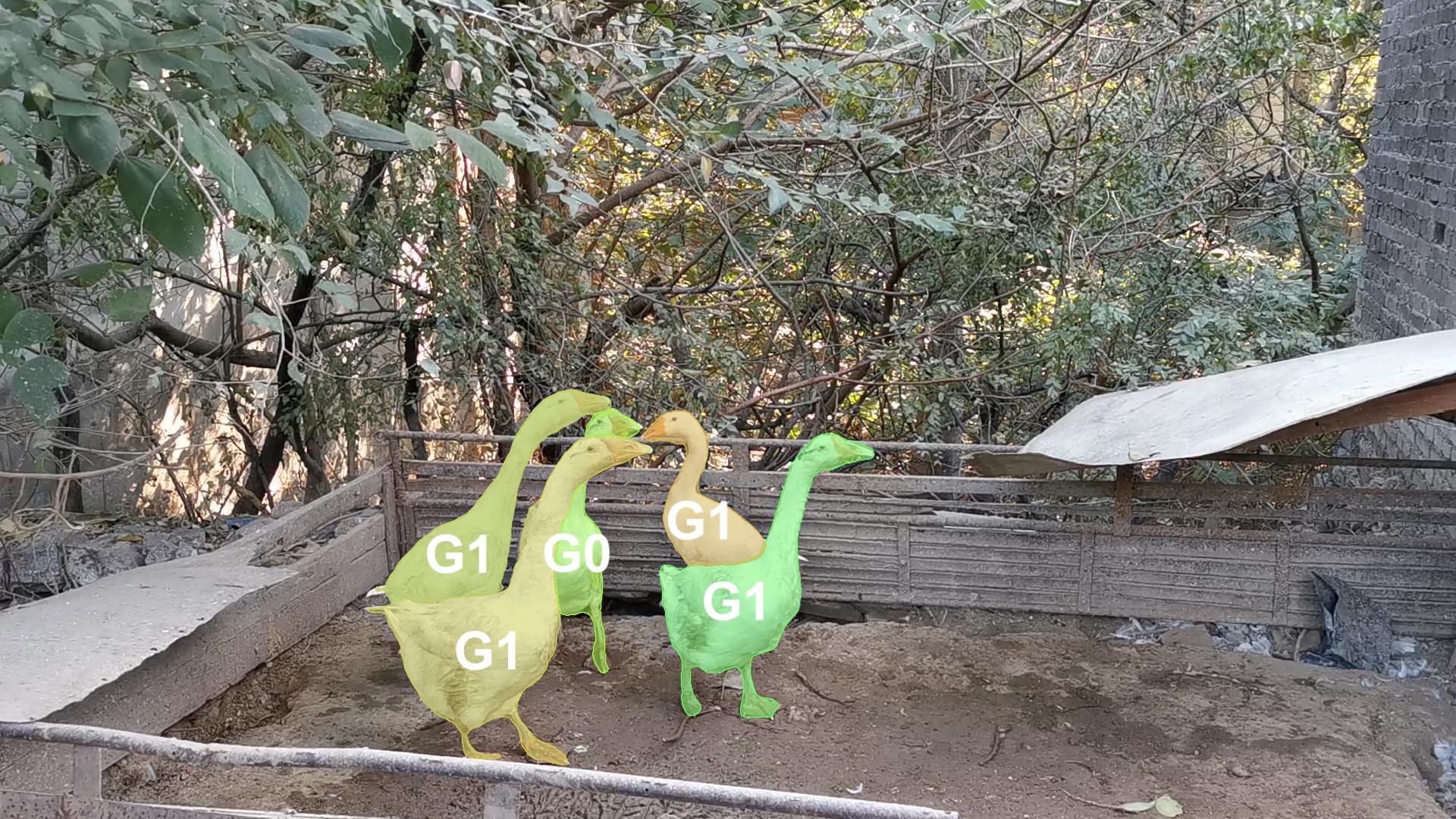}  \\
\includegraphics[width=.22\linewidth]{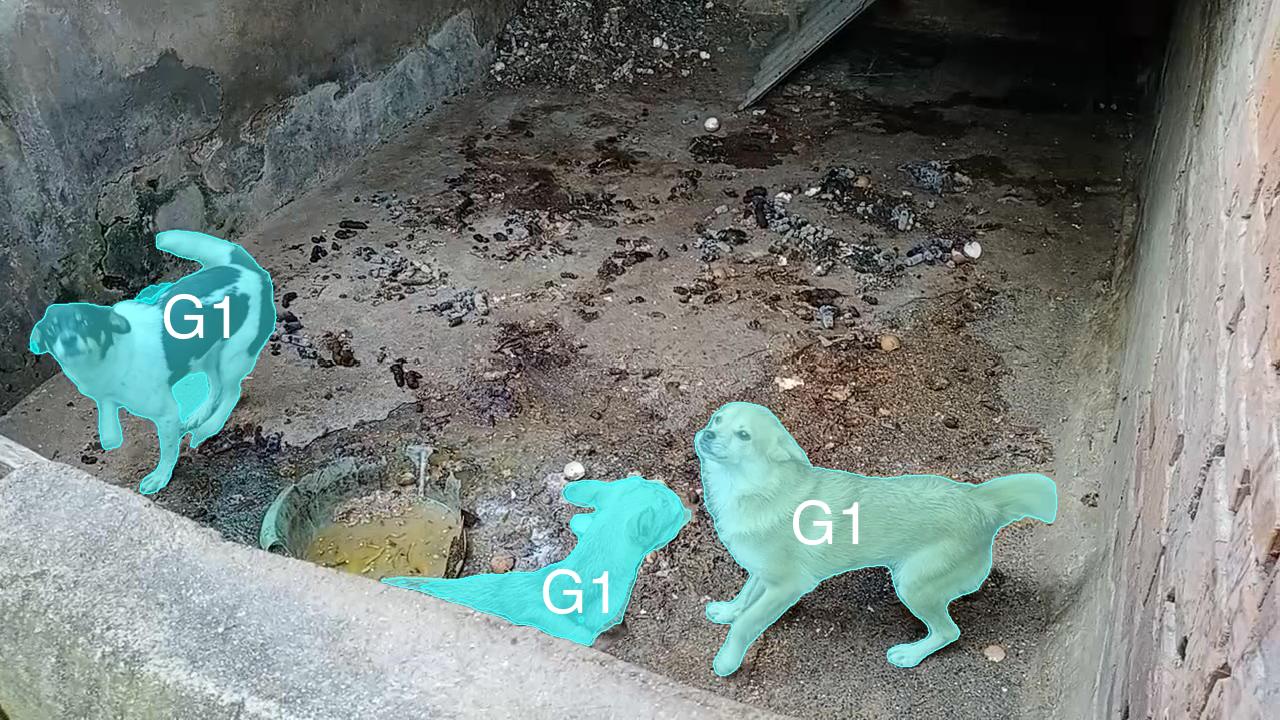} &  
\includegraphics[width=.22\linewidth]{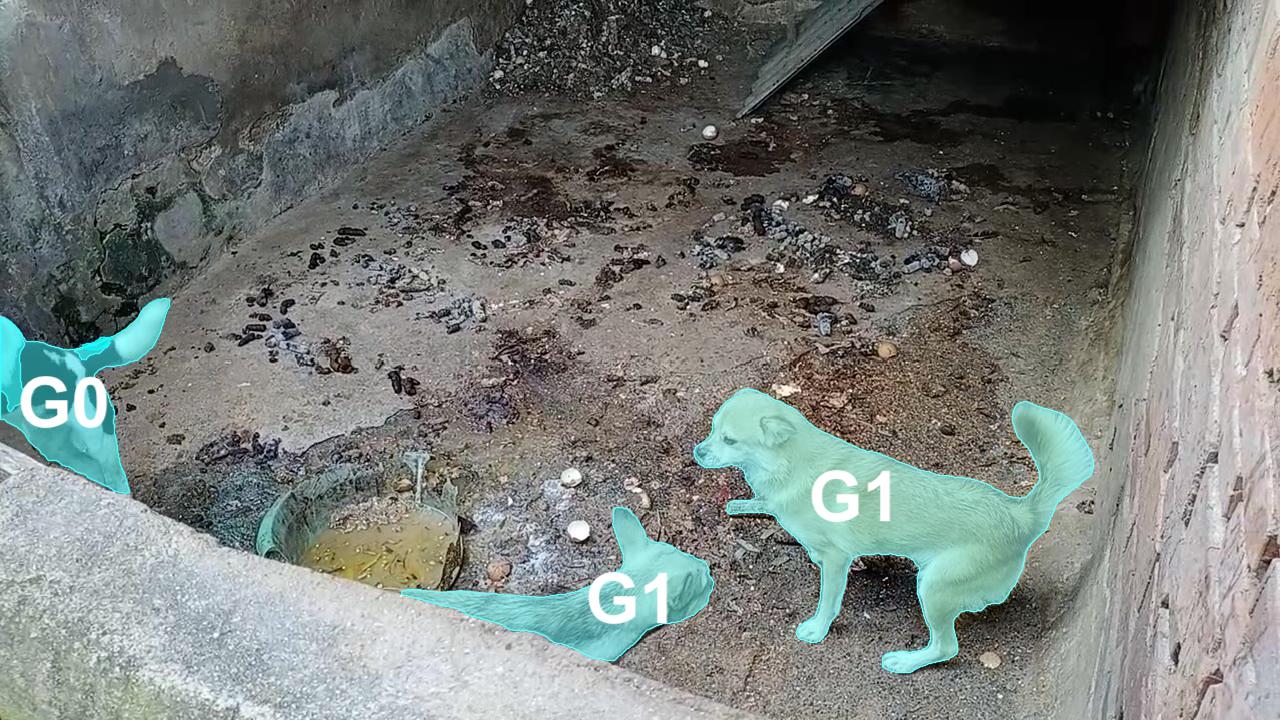} & 
\includegraphics[width=.22\linewidth]{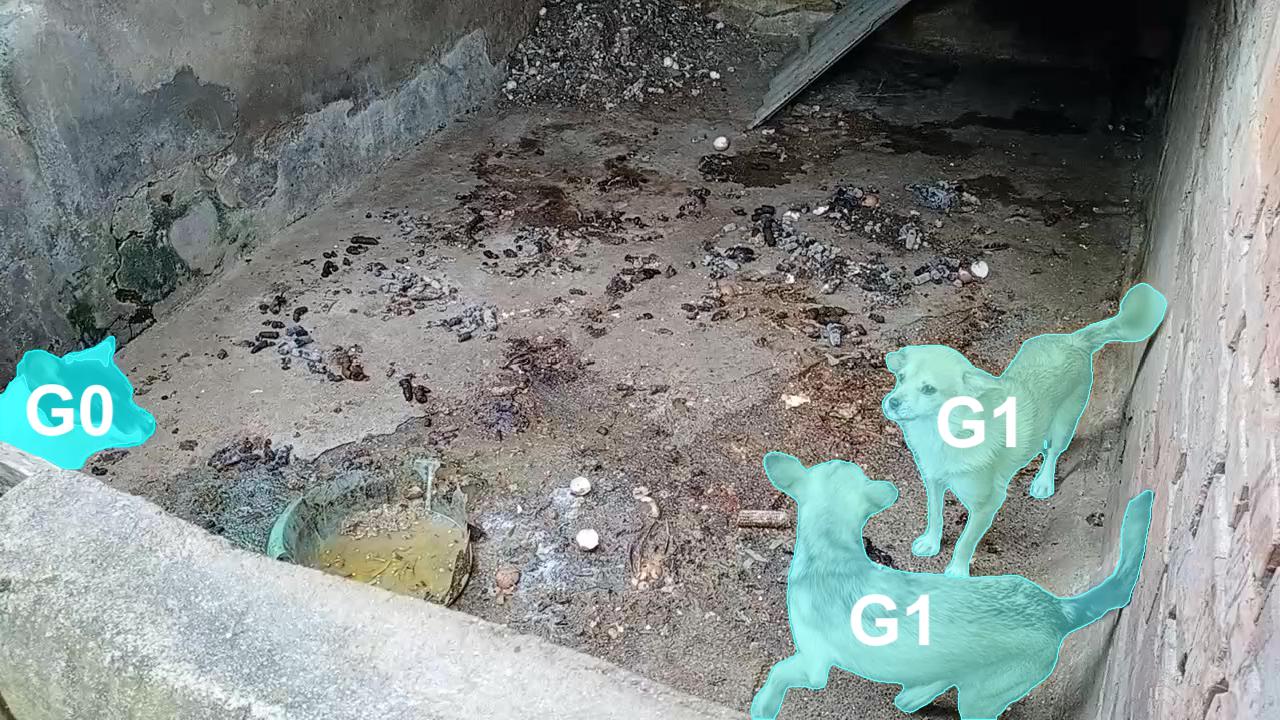} & 
\includegraphics[width=.22\linewidth]{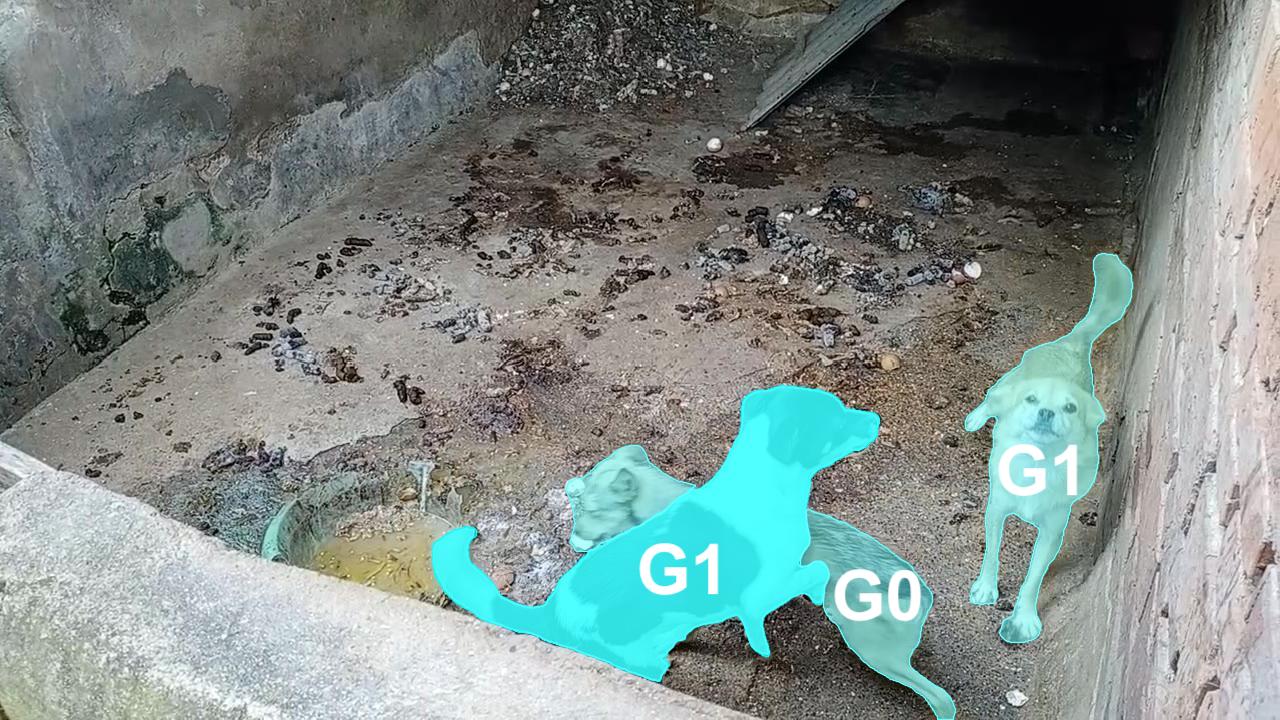}
\end{tabular}
\caption{Qualitative analysis of \textbf{\name{}} on complex \textbf{OVIS}-validation videos. We show severe occlusion and  scenarios with complicated instance dynamics. The overall as well as instance specific gate activation is presented with each video. Notably, with the presence of occlusion and abrupt changes, the Gate activation shows downward spike. For example, In the second video, we can see that the gate activation goes down significantly when the cars get occluded by the pedestrians. This characteristic phenomena is consistent across videos and various scenarios.}
\label{fig:qualitative}
\end{figure*}

\section{Qualitative Results}
Fig. \ref{fig:qualitative}, shows the predictive ability of \name{} on the highly challenging OVIS dataset, spanning a range of scenarios. These videos contain diverse situations, such as slight occlusion (1st row), a rapidly moving person (2nd row), ducks with similar appearances crossing each other's paths (3rd row), and severe occlusion featuring the disappearance and reappearance of a dog (4th row). Notably, \name{} consistently segments and tracks instances across these demanding circumstances.

\begin{figure}[t!]
    \centering
    \includegraphics[width=0.99\textwidth]{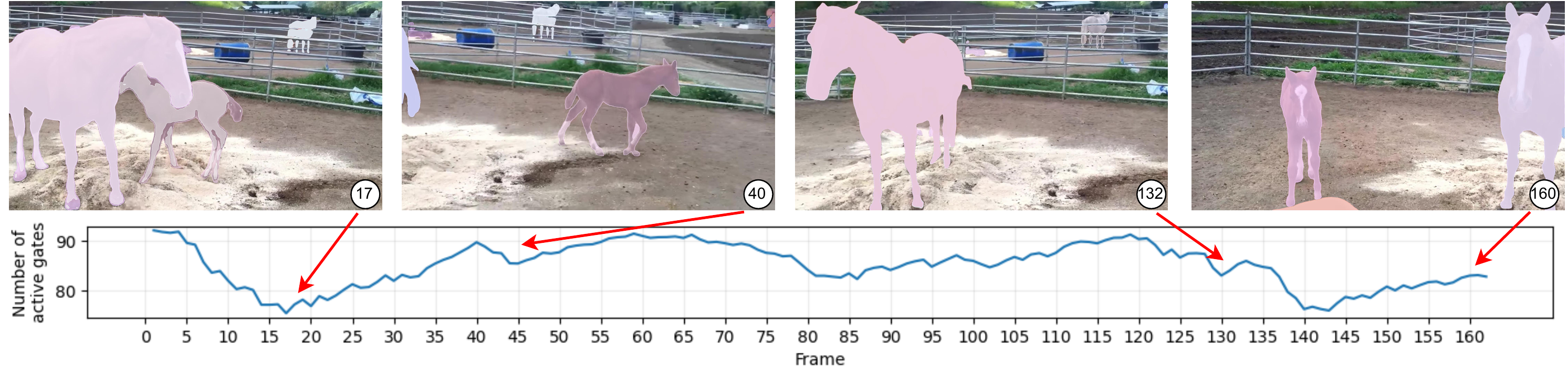}
    \caption{Number of activated gates across frames. Whenever the instance queries undergo shocks, in this example, because of disappearing and reappearing objects, the number of active gates drops. In the last displayed frame, many gates have turned active again, as the objects are clearly visible.}
    \label{fig:qual-active-gates}
\end{figure}

To gain deeper insights into our model's operational principles, we have visualized the output of the gating mechanism in Fig. \ref{fig:qualitative} and \ref{fig:qual-active-gates}. The gating signal in Fig \ref{fig:qualitative} is mostly \lq zero' when subjected to occlusion. Fig. \ref{fig:qual-active-gates} illustrates a decrease in active gates during occlusion or abrupt changes in the video. For example, gate activation is significantly reduced when the two horses start to occlude each other or when one of them vanishes. Similarly, the count drops again when the smaller horse goes through the severe occlusion. The last frame does not possess any complicated dynamics, and we observe an increasing trend of gate activation in that region of time. This highlights the model's ability to suppress noisy queries from the current frame during sudden changes. 

\subsection{Ablation Studies}
In this section, we present an ablation analysis of \name{} using our baseline models: InstanceFormer and GenVIS. Specifically, we investigate the impact of the GRAtt module on propagation.

As shown in Table \ref{tab:mem_gate_ablation}, our ablation study highlights the superiority of \name{} over two recently proposed architectures, InstanceFormer and GenVIS, on the challenging  OVIS dataset. We compare two versions of both InstanceFormer and GenVIS, with and without memory. The memory-free versions serve as our baselines, upon which we built \name{}-I and \name{}-G. Note that in both InstanceFormer and GenVIS, we only replaced their decoder with our GRAtt module, keeping all other components as original.

Compared to InstanceFormer, \name{}-I demonstrated a significant improvement, outperforming the memory-free baseline by $5.2\%$ in terms of Average Precision (AP). Furthermore, it surpassed the memory-inclusive InstanceFormer by $2.3\%$ AP. Similarly, compared to GenVIS, \name{}-G also performs better than the corresponding baseline. It improves the performance of the memory-free GenVIS by $1.2\%$ AP and even surpasses the memory-inclusive GenVIS by $0.4\%$ AP. Importantly, the performance improvement complemented by the reduction of GFLOPs of about $37.8\%$ indicates a better operating point in the efficiency vs. performance landscape. The stream of past instances through the residual connection serves as a dynamically allocated memory, thus replacing the conventional memory bank. Moreover, the GRAtt decoder converges $33\%$ faster than the baselines. 
 
\begin{table*}[h]
\centering
\resizebox{0.85\linewidth}{!}
{
        \begin{tabular}{c|c|c|ccccc|cc}
        \toprule 
         Method & Mem & \textbf{GRAtt}&$\mathrm{AP}$ & $\mathrm{AP}_{50}$ & $\mathrm{AP}_{75}$ & $\mathrm{AR}_1$ & $\mathrm{AR}_{10}$&GFLOPs  \\
        \hline \hline 
        InstanceFormer &   &  &17.1 & 35.5 & 15.4 & 9.8 & 24.7 & -\\
        InstanceFormer & \cmark &   & 20.0 & 40.7 & 18.1 & 12.0 & 27.1 & -\\
         \name{}-I   &  & \cmark &\textbf{22.3} & \textbf{43.0} & \textbf{19.5} & \textbf{12.1} & \textbf{29.8} & - \\
        \midrule
        
        GenVIS &  &  & 35.0 & 57.6 & 36.3 & 16.3 & 36.3 & 2.4\\
        GenVIS & \cmark & & 35.8 & \textbf{60.8} & 36.2 & 16.3 & 39.6 & 3.7\\
        \name{}-G &    & \cmark& \textbf{36.2} & \textbf{60.8} & \textbf{36.8} & \textbf{16.8} & \textbf{40.0} &\textbf{2.3}\\
        \bottomrule
        \end{tabular}
    }
    \caption{Impact of our GRAtt decoder when integrated with two propagation-based VIS Frameworks. It improves on both the memory-free and memory-based variants and, at the same time, reduces overall computation of GFLOPs (less is better) and increases speed.}
    \label{tab:mem_gate_ablation}
\end{table*}

\section{Limitation and Future Work}
\name{} explicitly models the subproblems of VIS, i.e., short, and long-term occlusion, instance appearance and disappearance, and rectification from abrupt noise. However, we observe that crossover trajectories still suffer from ID Switching. We attribute this to the lack of explicit trajectory modeling. \name{} relies on query propagation for implicit trajectory modeling. We hope to extend the GRAtt decoder to model the optimal trajectory.

\section{Conclusion}
In this work, we have introduced the GRAtt decoder, a novel VIS module with a self-rectifying mechanism capable of absorbing disturbance in video frames and recovering degraded instance features. The self-rectifying mechanism is learned purely based on data without any heuristic to find an optimum discrete decision between relying on the current frame or falling back to the past one. It can be seamlessly integrated into the family of query propagation-based methods. It simplifies the VIS pipeline by eliminating the need for a memory bank and, at the same time, sparsifies the attention space. We hope our work will accelerate research toward sparse and efficient video processing. This would enhance video instance segmentation and have significant implications for video analytics.

\label{sec:concl}
\newpage
\bibliographystyle{plainnat}
\bibliography{references}

\clearpage
\appendix

\section{Detailed Ablation for \name{}}
This section provides ablation studies focusing on various design choices of the gated residual connection and attention mask configuration. Unless otherwise stated, all the experiments are done on the OVIS~\cite{OVIS} dataset.

\subsection{Placement of Residual Connection}
\begin{figure}[H]
    \centering
    \includegraphics[width=1\linewidth]{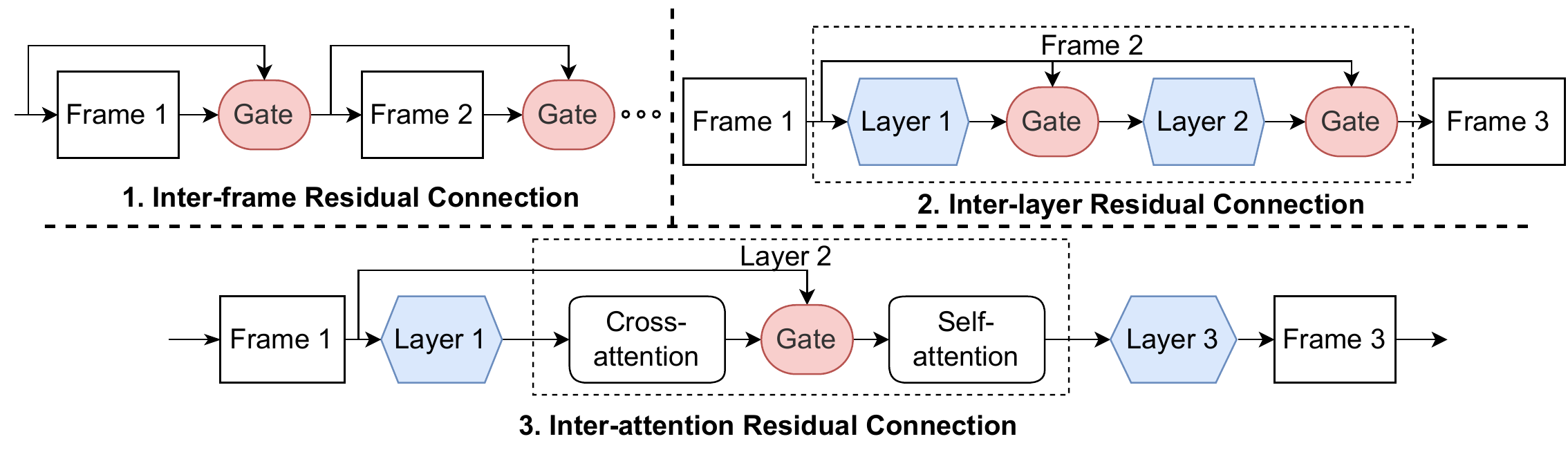}
    \caption{\textbf{Gated Residual Connection}. The gate can be placed between frames, layers of decoder, or inside the layer. \textbf{\name{}} incorporates the inter-attention gating mechanism.}
    \label{fig:rescon}
\end{figure}

Figure \ref{fig:rescon} presents different placement options for the gated residual connection. In the first configuration, the connection is positioned between consecutive frames, allowing the current query to propagate based on its relevance after processing the entire frame. 

In the second configuration, the connection is placed between decoder layers. This provides greater flexibility to the network, enabling it to independently assess the relevance of instance queries and their propagation at each layer. 

The final configuration involves placing the connection inside each layer between its cross- and self-attention, which is our default setting. 

Table \ref{tab:my_label1} shows that the inter-frame residual connection has a detrimental impact on the overall performance, while the inter-layer connection improves it. This suggests that the inter-layer connection enhances the network's flexibility, as each layer contributes to the gating process, resulting in improved performance. 

Finally, our default inter-attention residual connection further enhances performance by introducing even more flexibility and rectifying noisy queries before self-attention. Hence, from Table \ref{tab:my_label1}, it is evident that a more flexible gated residual connection directly translates to better performance.
\begin{table}[H]
    \centering
        \begin{tabular}{c|ccccc}
        \hline 
         Gate Placement & $\mathrm{AP}$ & $\mathrm{AP}_{50}$ & $\mathrm{AP}_{75}$ & $\mathrm{AR}_1$ & $\mathrm{AR}_{10}$ \\
        \hline \hline 
        Inter-Frame& 34.7 & 57.7 & 35.3 & 16.4 & 38.2 \\
        Inter-Layer& 35.0 & 57.6 & \textbf{36.3} & 16.3 & 36.3 \\
        Inter-Attention (ours)& \textbf{35.2} & \textbf{58.5} & 35.8 & \textbf{16.5} & \textbf{39.0} \\
        \hline
        \end{tabular}
    \caption{ Performance comparison with different placements of gated residual connection. No attention mask is used in this experiment. }
    \label{tab:my_label1}
\end{table}

\subsection{Attention Mask Configurations}
\begin{figure*}[!h]
    \centering
    \includegraphics[width=1\linewidth]{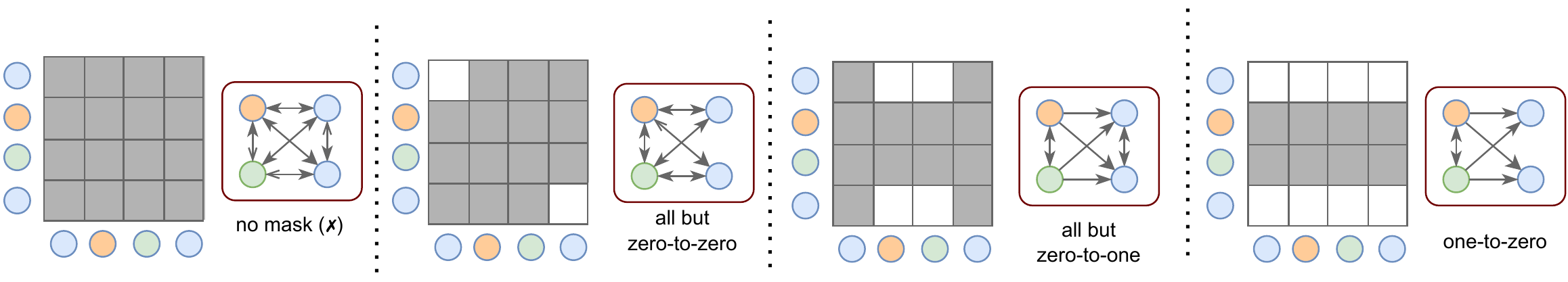}
    \caption{Various attention schema are illustrated here. Each comprises of a mask and a graphical representation. Note that, for simplification, the self-directing arrows in the attention graph have been omitted. The white cells represent the query pairs that have been masked. The one on the left is a typical self-attention. In the second scenario, the Zero gated queries (colored blue) do not interact with each other. In the 3rd one, they do not attend to One Gated Queries. Finally, the rightmost one is \textbf{GRAtt}-mask, which excludes Zero gated queries from the attention query set. }
    \label{fig:attn}
\end{figure*}
We take the gate outcome after the cross-attention to generate a self-attention mask. We experimented with three variations of the mask as illustrated in Fig. \ref{fig:attn}. The leftmost \lq all-to-all' figure corresponds to a standard self-attention where all instance queries attend to each other. Then \lq all but zero-to-zero' restricts the irrelevant object queries with a \lq zero' gate output to attend to each other. According to \lq all but zero-to-one', the irrelevant queries do not attend to the relevant queries with a gate value of \lq one.' Finally, \lq one-to-zero' excludes the irrelevant instances from the query set of self-attention. For all the variants, the queries with gate activation value \lq one' attend to the global key set. Specifically, all except the final variant allow the total or partial update of the irrelevant queries resulting in a sub-optimal performance as reported in Tab. \ref{tab:my_label2}. 

\subsection{Qualitative Analysis}
In this section, we visualize query features across video frames through t-SNE plots. Additionally, we include more predicted visualizations of GRAtt-vis for OVIS~\cite{OVIS}, and YTVIS-22~\cite{YouTube-VIS-2022} datasets.
\begin{figure*}[!h]
\centering
\includegraphics[width=\linewidth]{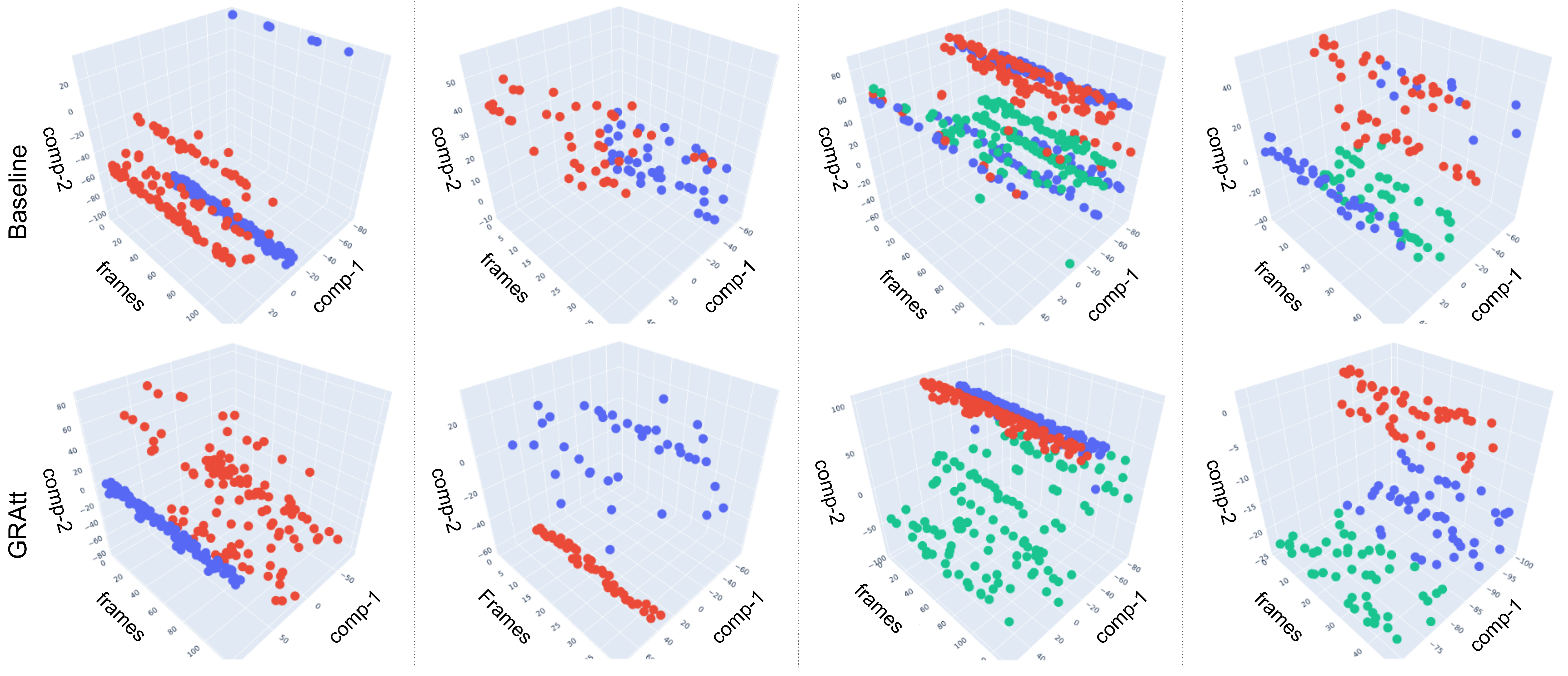}
\caption{Comparison of t-SNE embeddings between our Baseline model and GRAtt-VIS. The top row corresponds to the Baseline model w/o memory or gate. Each column depicts the t-SNE embeddings of predicted instances from one video. Colors differentiate the instances. GRAtt produces temporally more consistent instance features essential for tracking.}
\label{fig:qualitative_yt22}
\end{figure*}

\begin{table*}[!t]
    \centering
        \begin{tabular}{c|ccccc}
        \hline 
         Attn-Mask & $\mathrm{AP}$ & $\mathrm{AP}_{50}$ & $\mathrm{AP}_{75}$ & $\mathrm{AR}_1$ & $\mathrm{AR}_{10}$ \\
        \hline \hline 
        all but zero-to-zero &  33.9 & 58.0 & 34.5 & 16.4 & 37.7 \\
        all but zero-to-one & 34.7 & 57.1 & 35.3 & 16.3 & 39.4 \\
        all-to-all& 35.2 & 58.5 & 35.8 & 16.5 & 39.0 \\
        one-to-zero (ours) &  \textbf{36.2} & \textbf{60.8} & \textbf{36.8} & \textbf{16.8} & \textbf{40.0} \\
       
        \hline
        \end{tabular}
    \caption{ Performance comparison with different attention masks in the GRAtt decoder. We use our default Inter-Attention Residual Connection in this ablation.}
    \label{tab:my_label2}
\end{table*}

\begin{figure*}[!h]
\centering
\begin{tabular}{cccc}
\includegraphics[width=.22\linewidth]{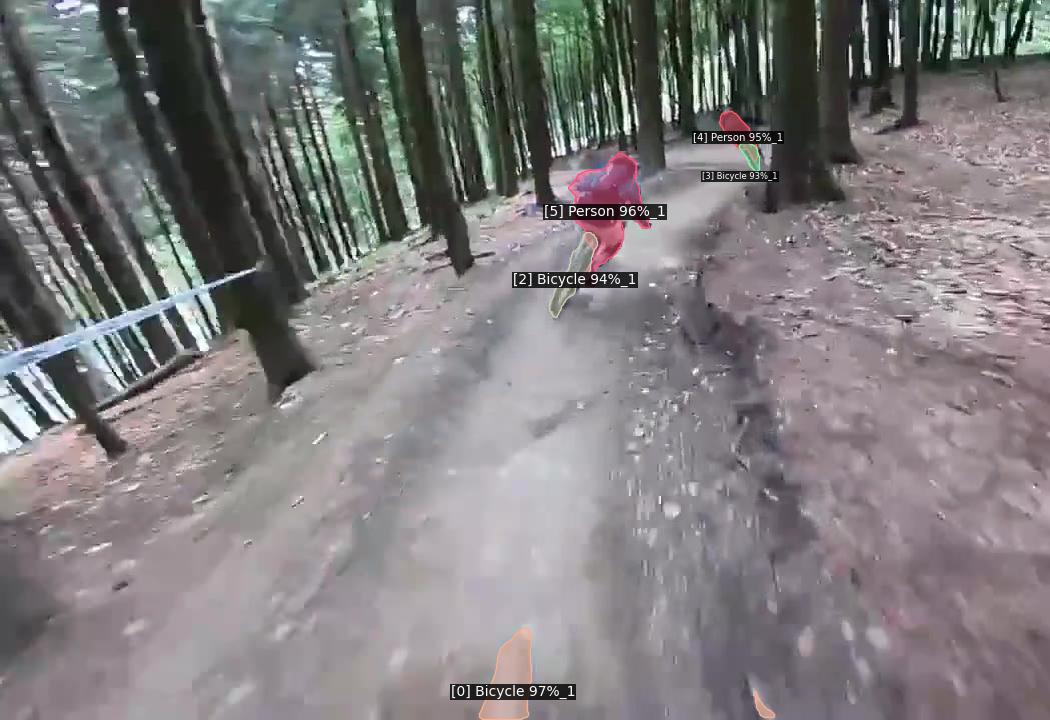} &
\includegraphics[width=.22\linewidth]{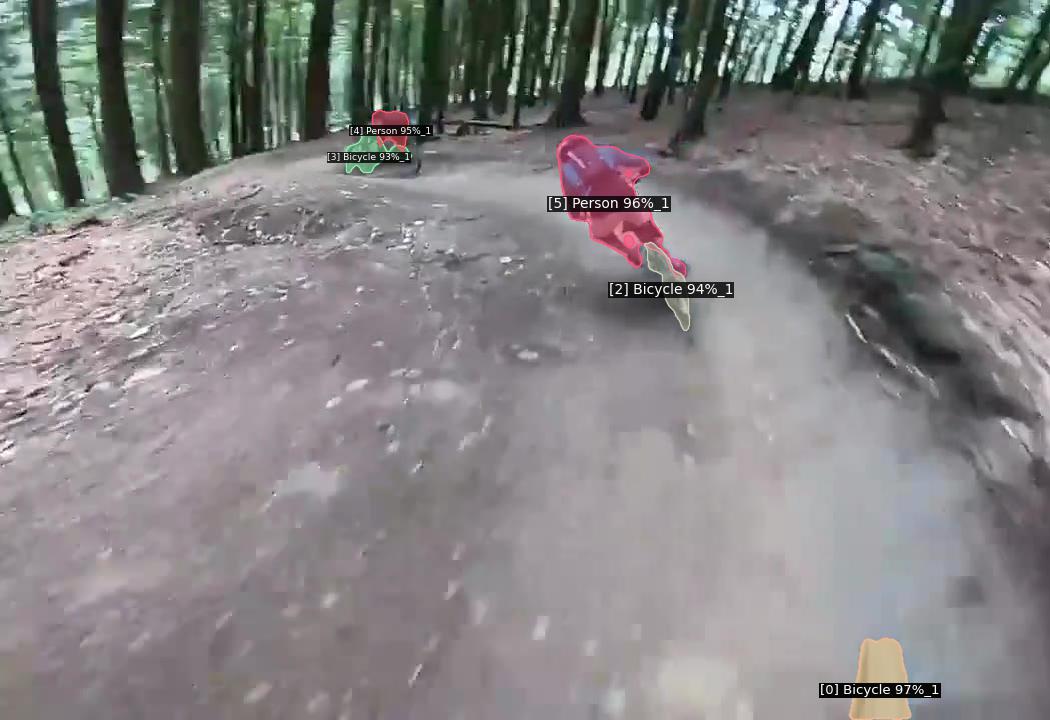} &
\includegraphics[width=.22\linewidth]{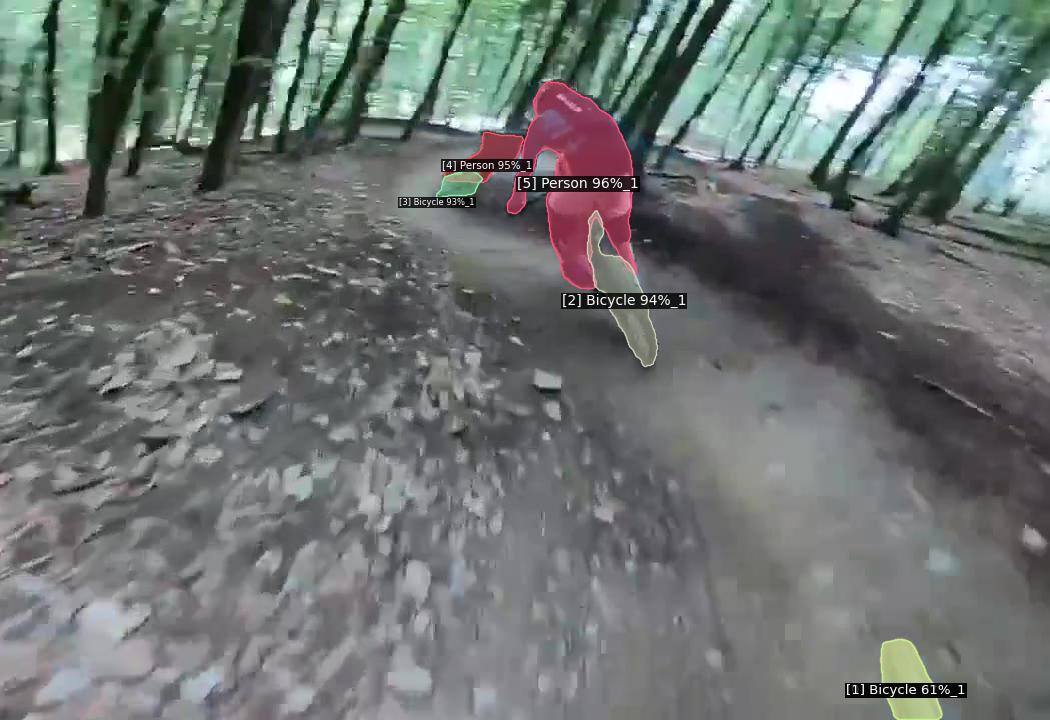} &
\includegraphics[width=.22\linewidth]{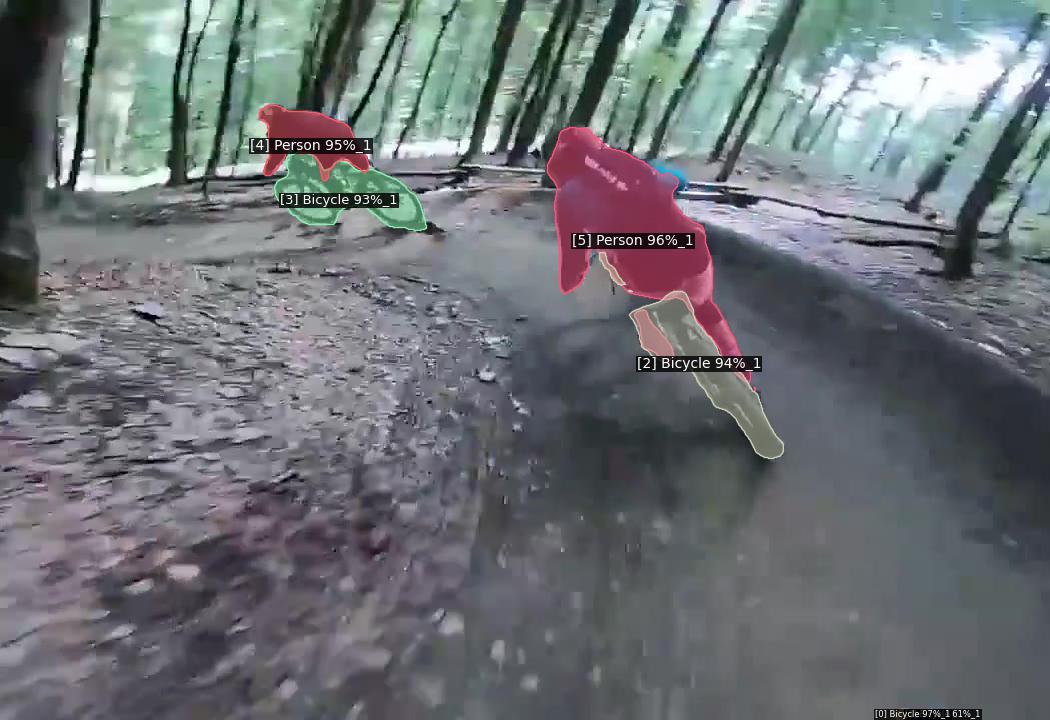} \\

\includegraphics[width=.22\linewidth]{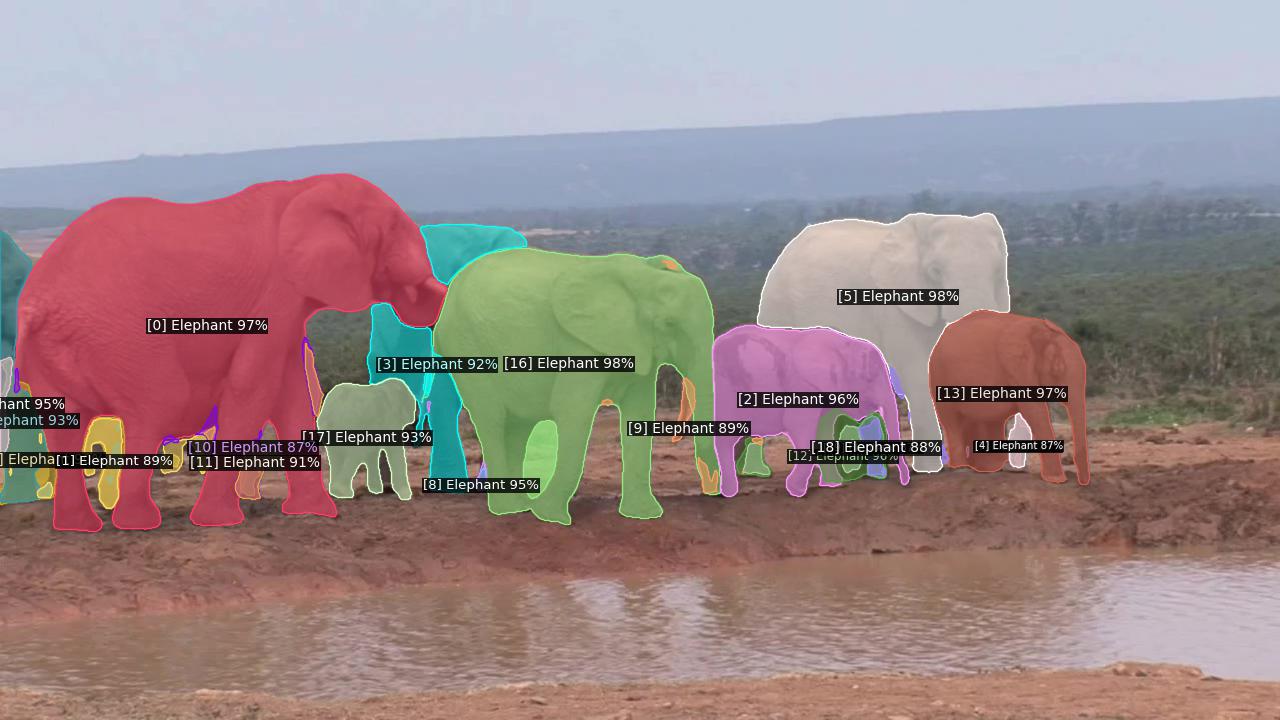} &
\includegraphics[width=.22\linewidth]{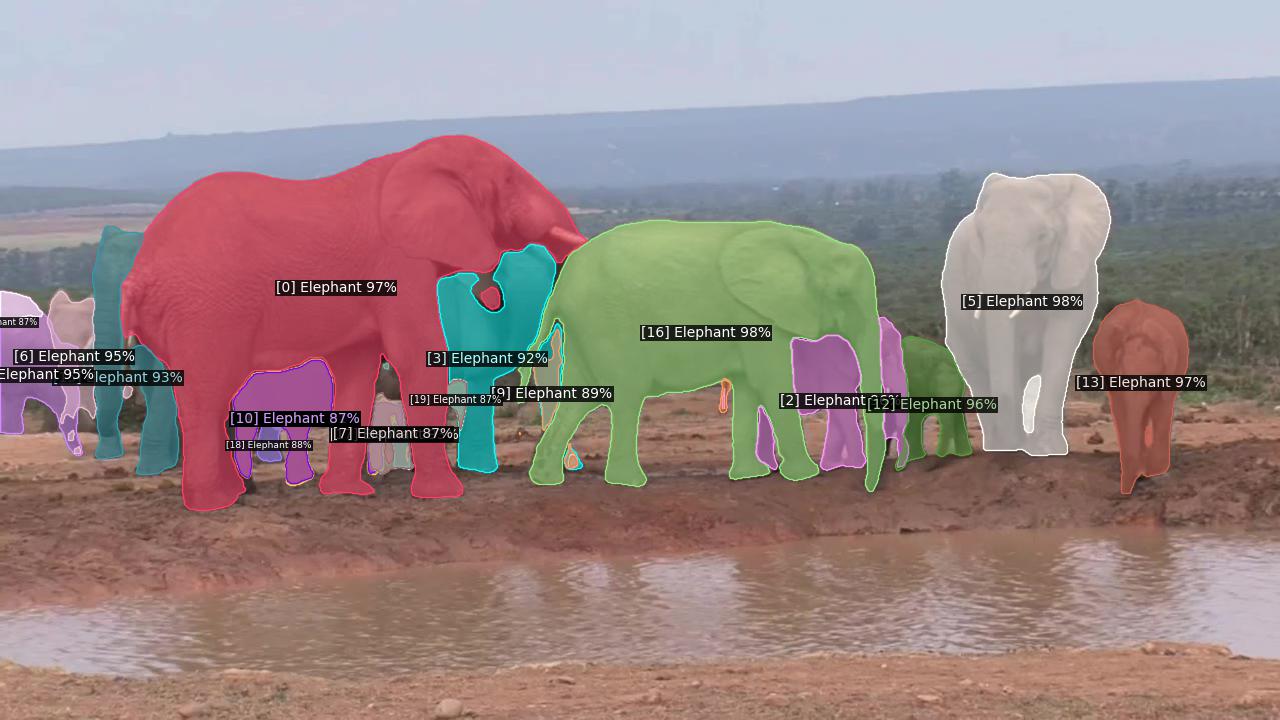} &
\includegraphics[width=.22\linewidth]{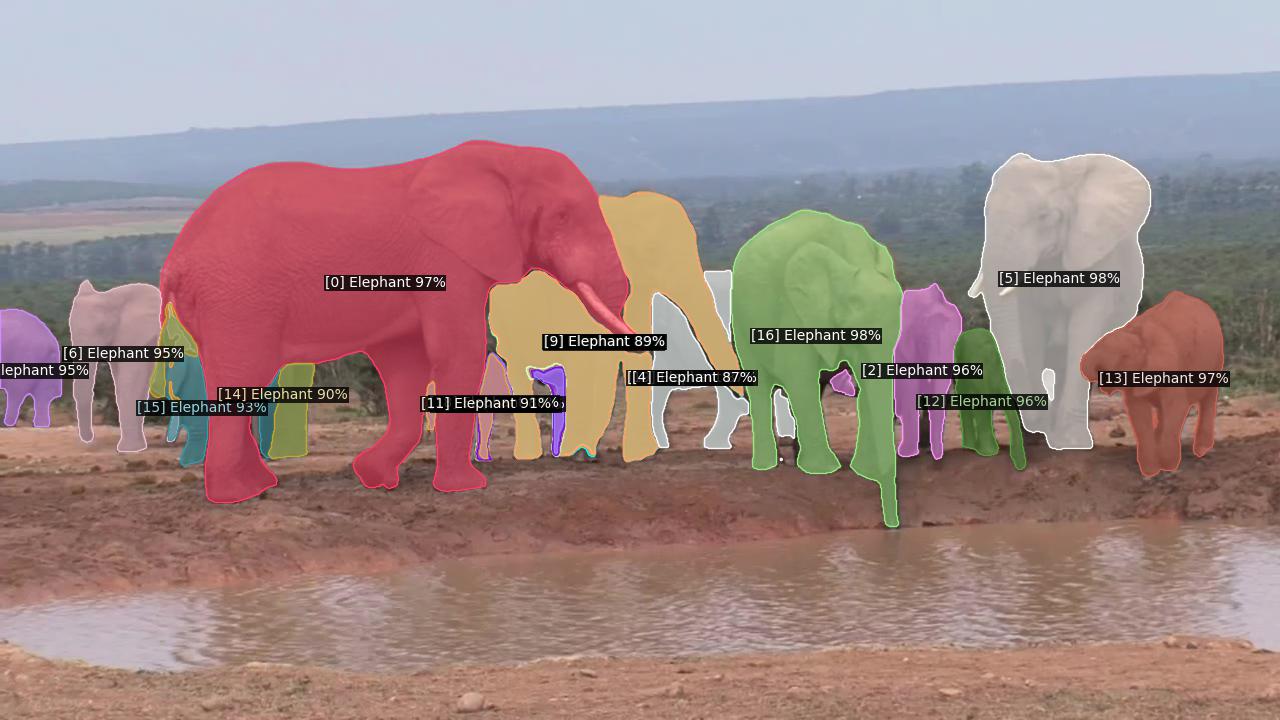} &
\includegraphics[width=.22\linewidth]{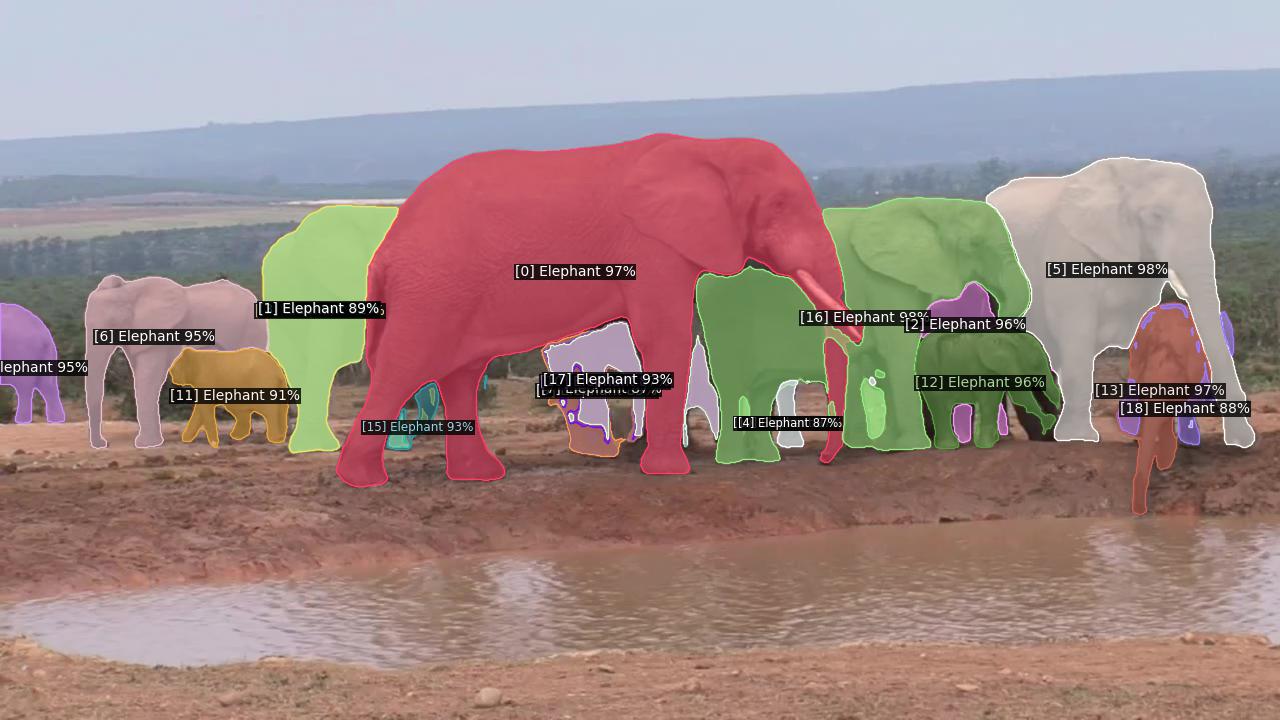} \\
\includegraphics[width=.22\linewidth]{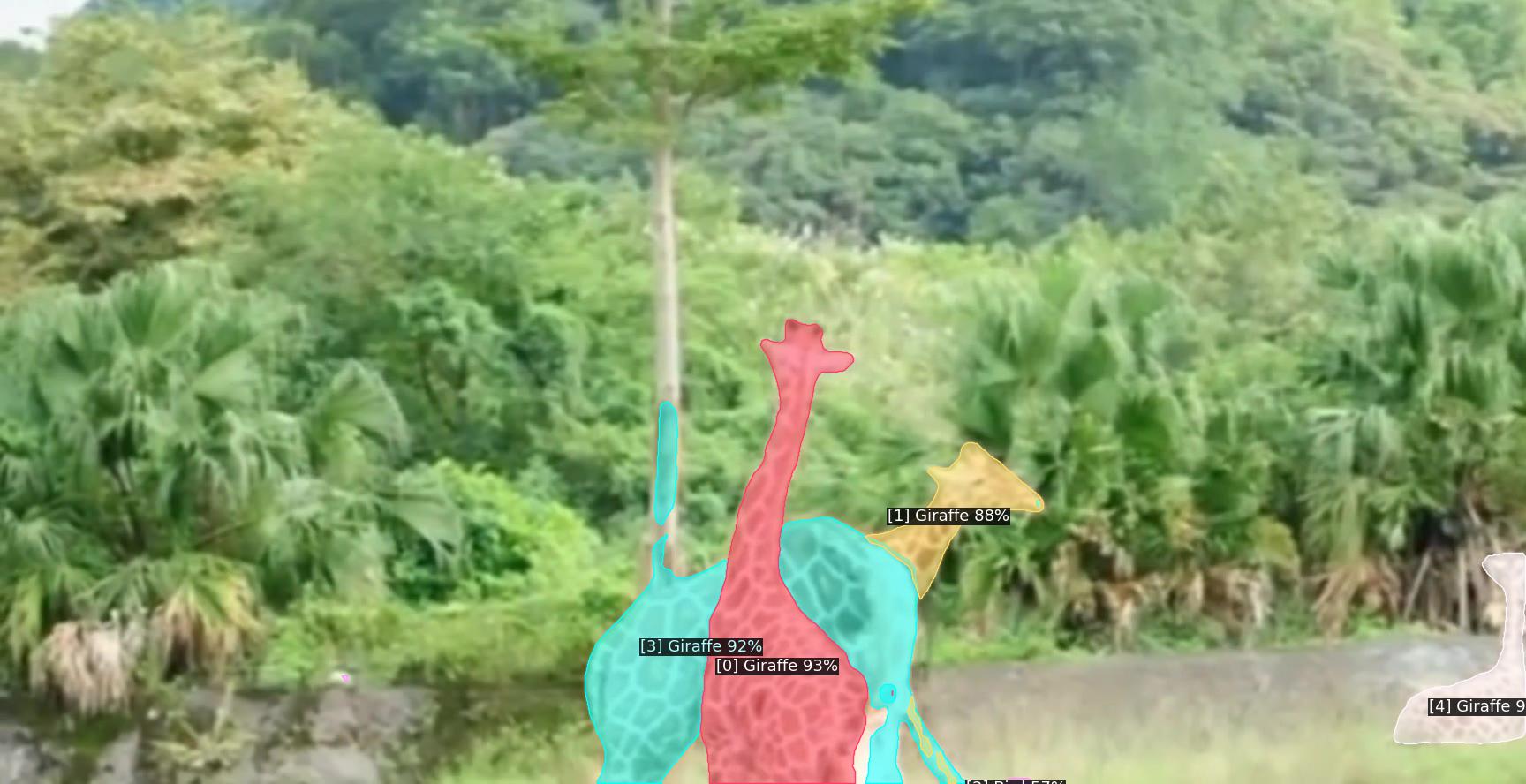} &
\includegraphics[width=.22\linewidth]{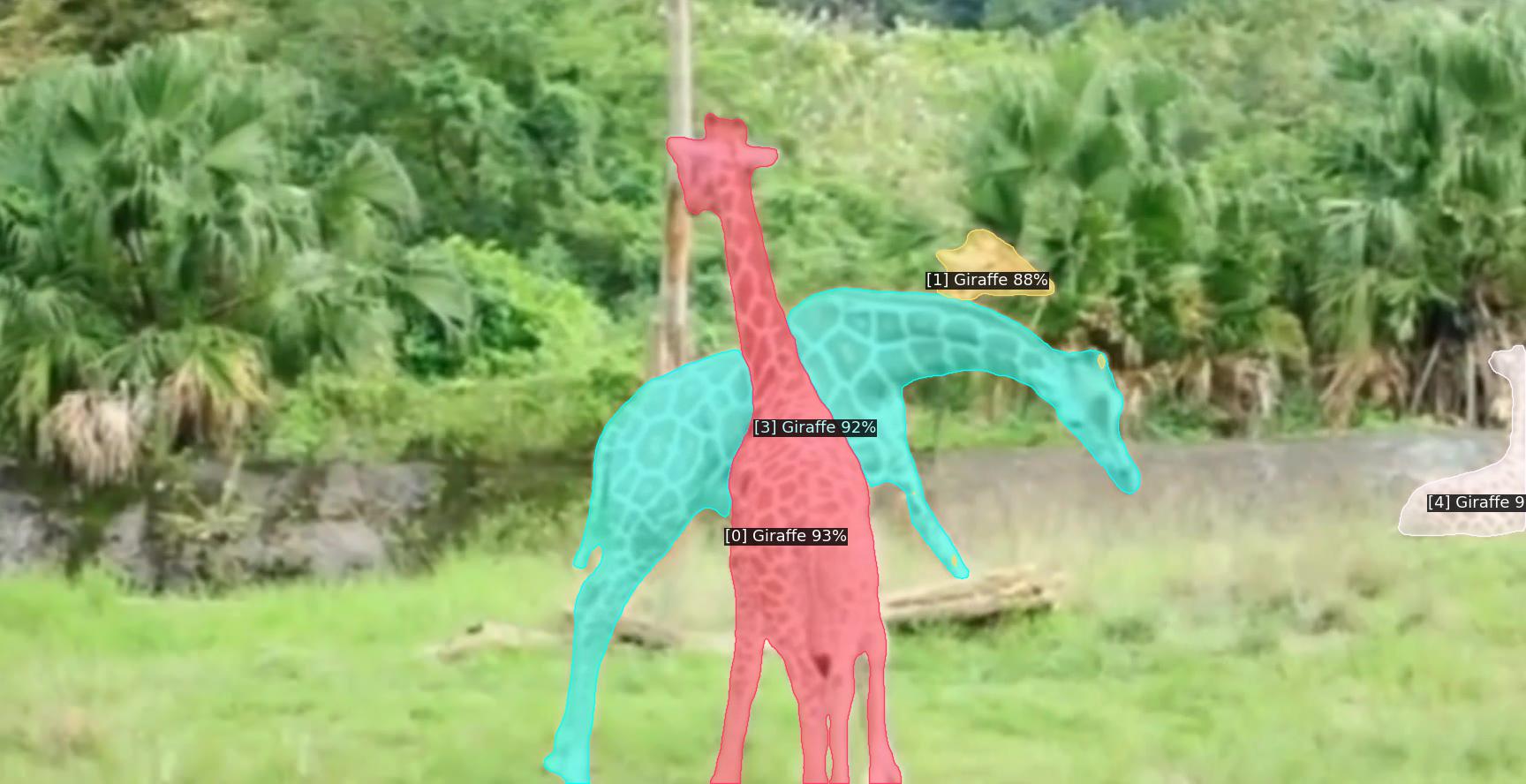} &
\includegraphics[width=.22\linewidth]{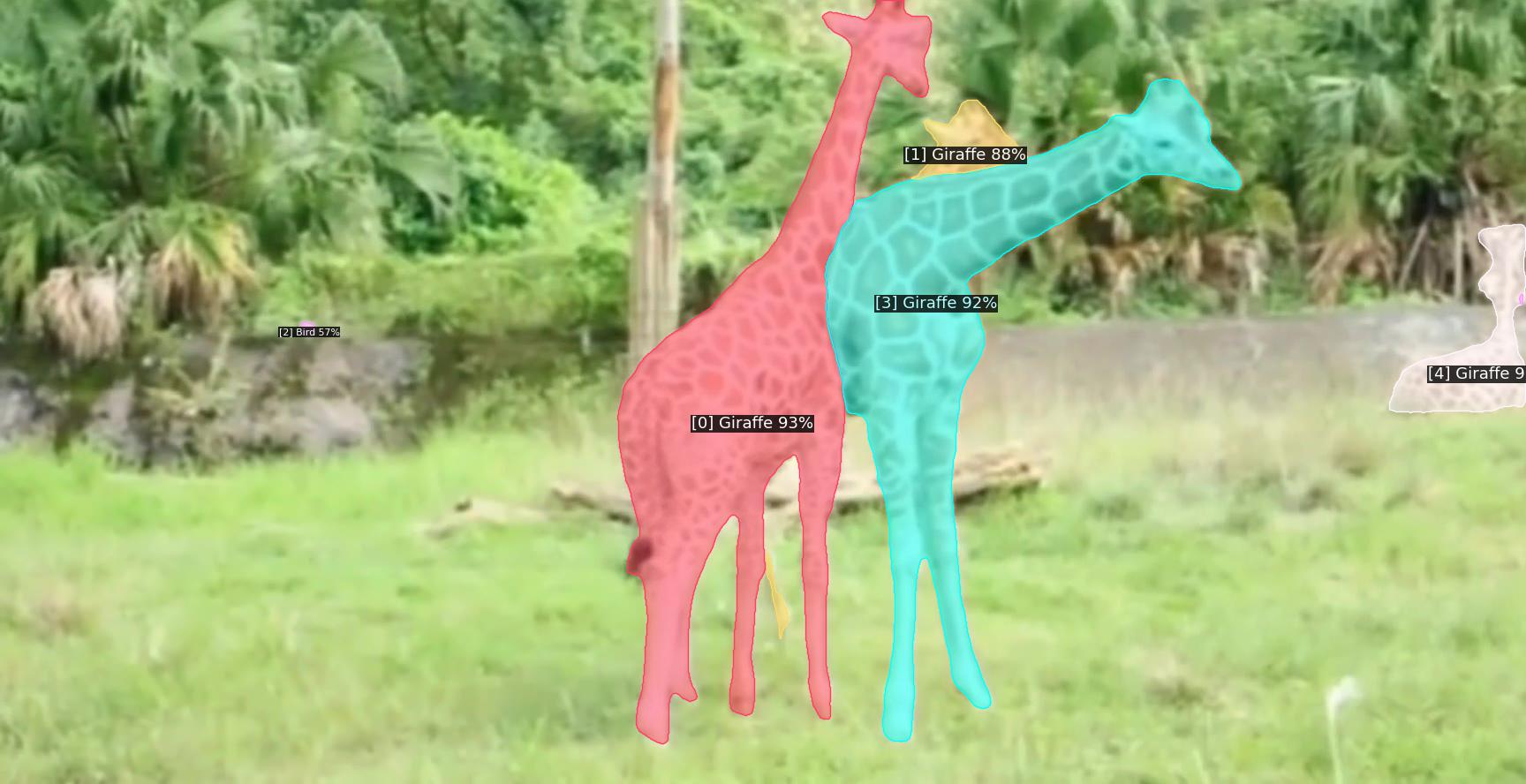} &
\includegraphics[width=.22\linewidth]{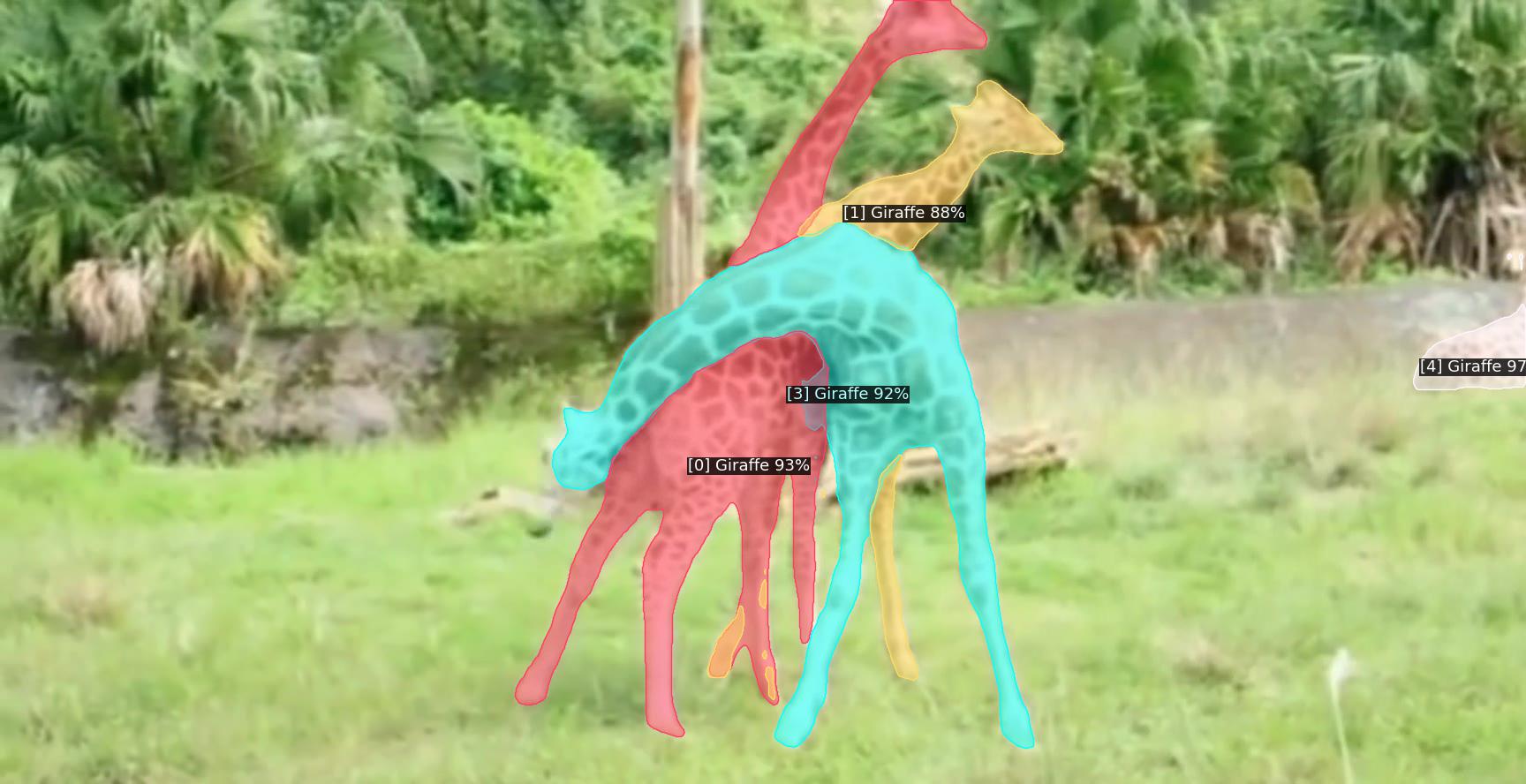} \\
\includegraphics[width=.22\linewidth]{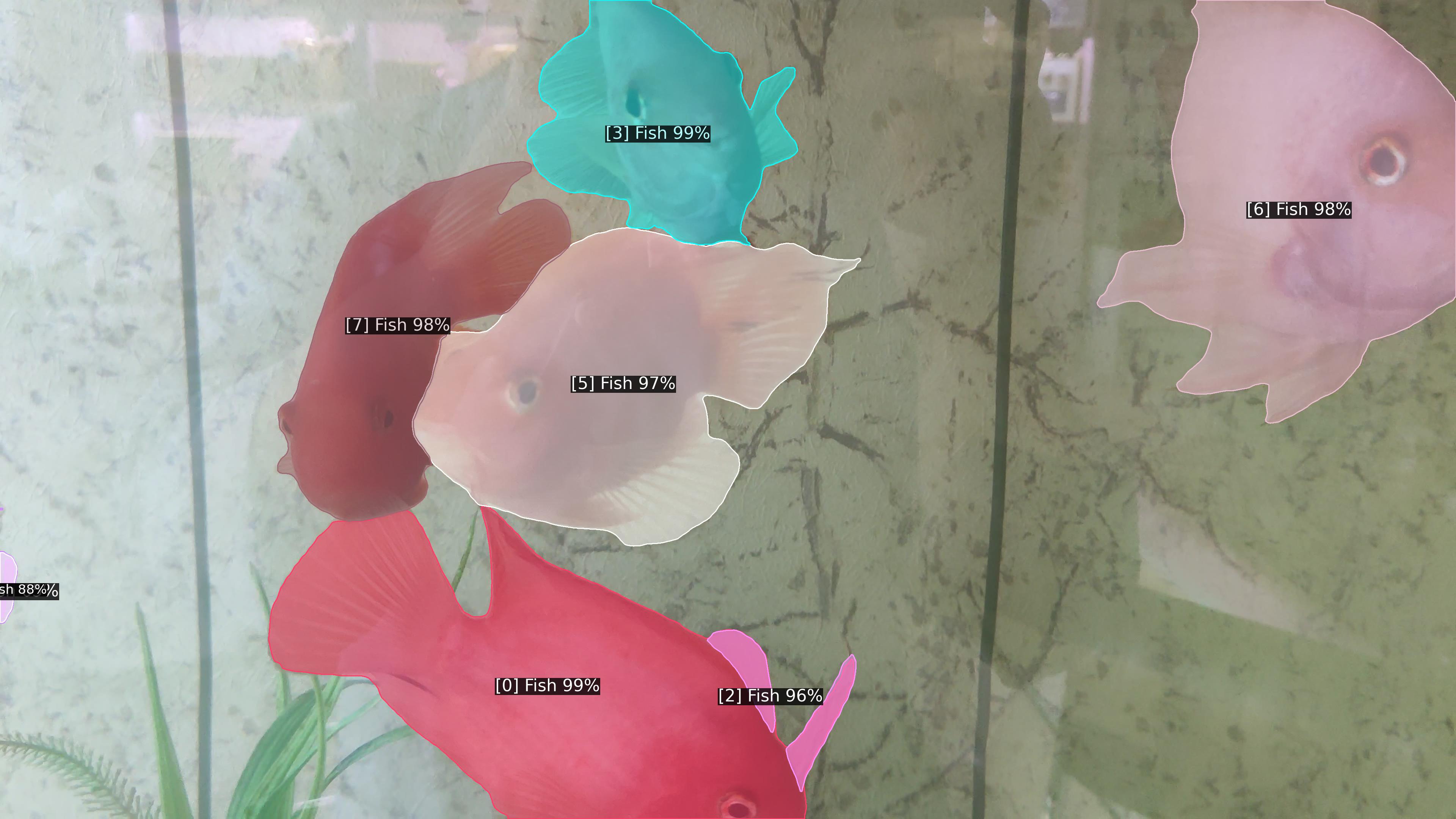} &
\includegraphics[width=.22\linewidth]{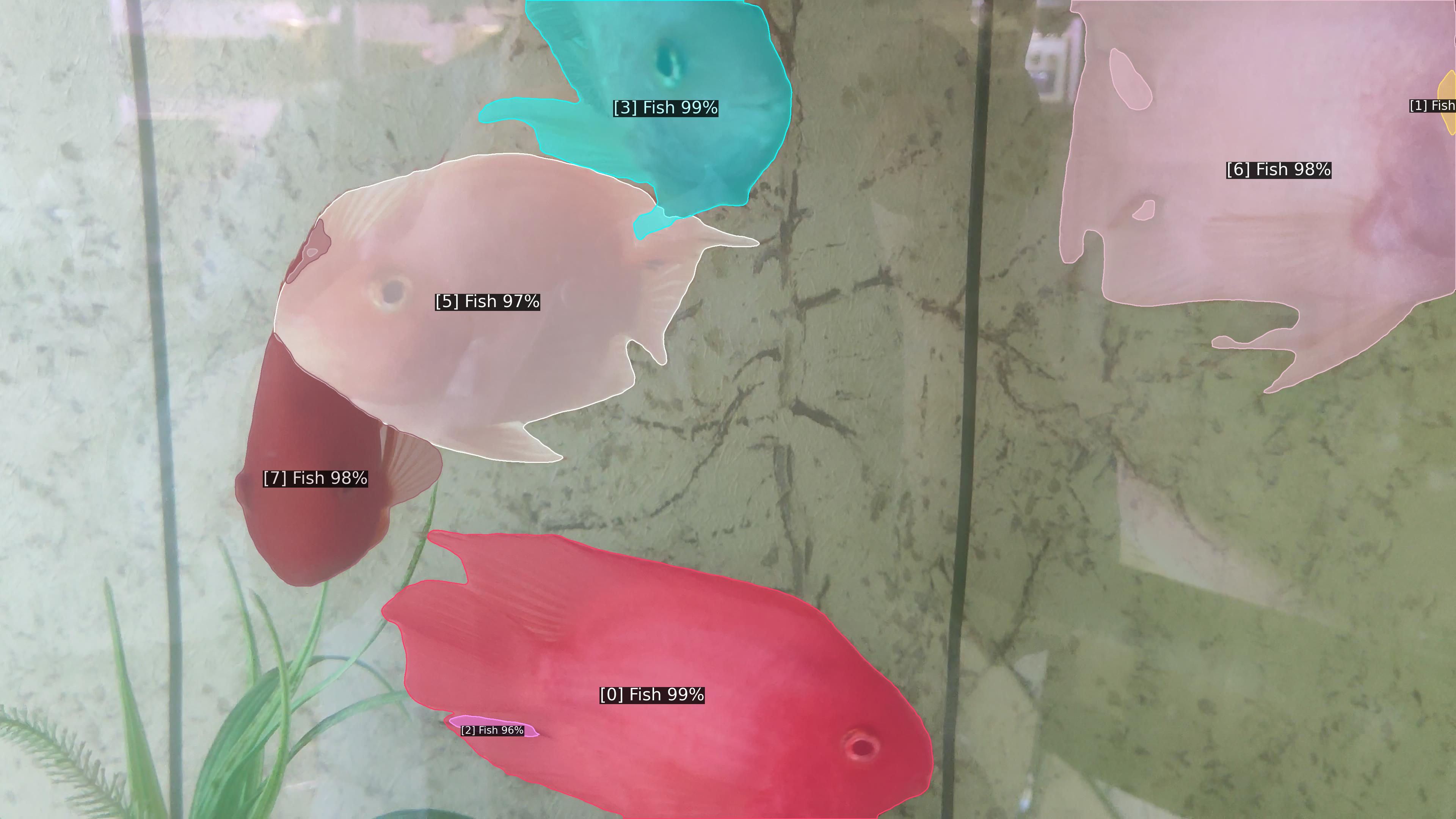} &
\includegraphics[width=.22\linewidth]{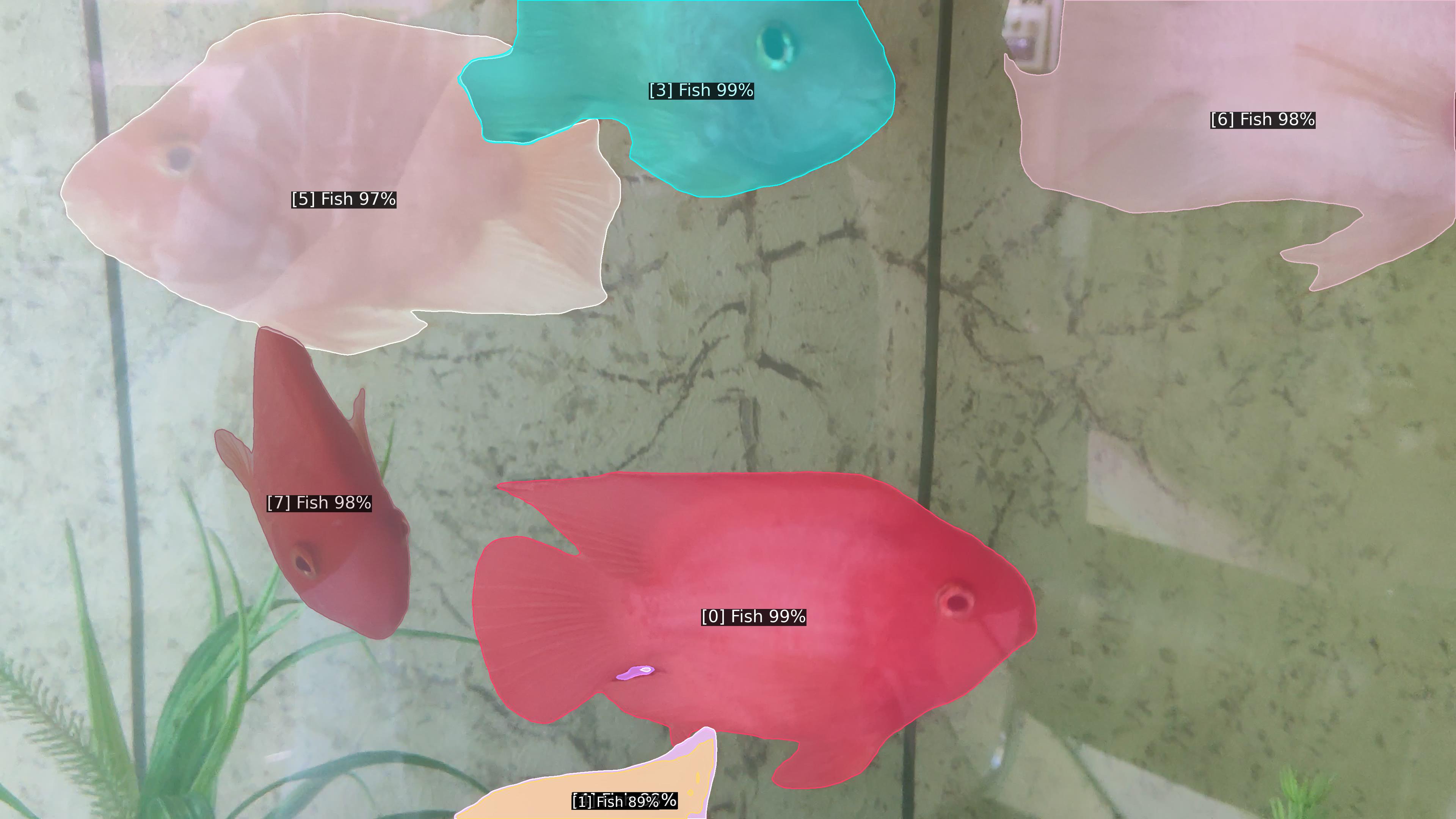} &
\includegraphics[width=.22\linewidth]{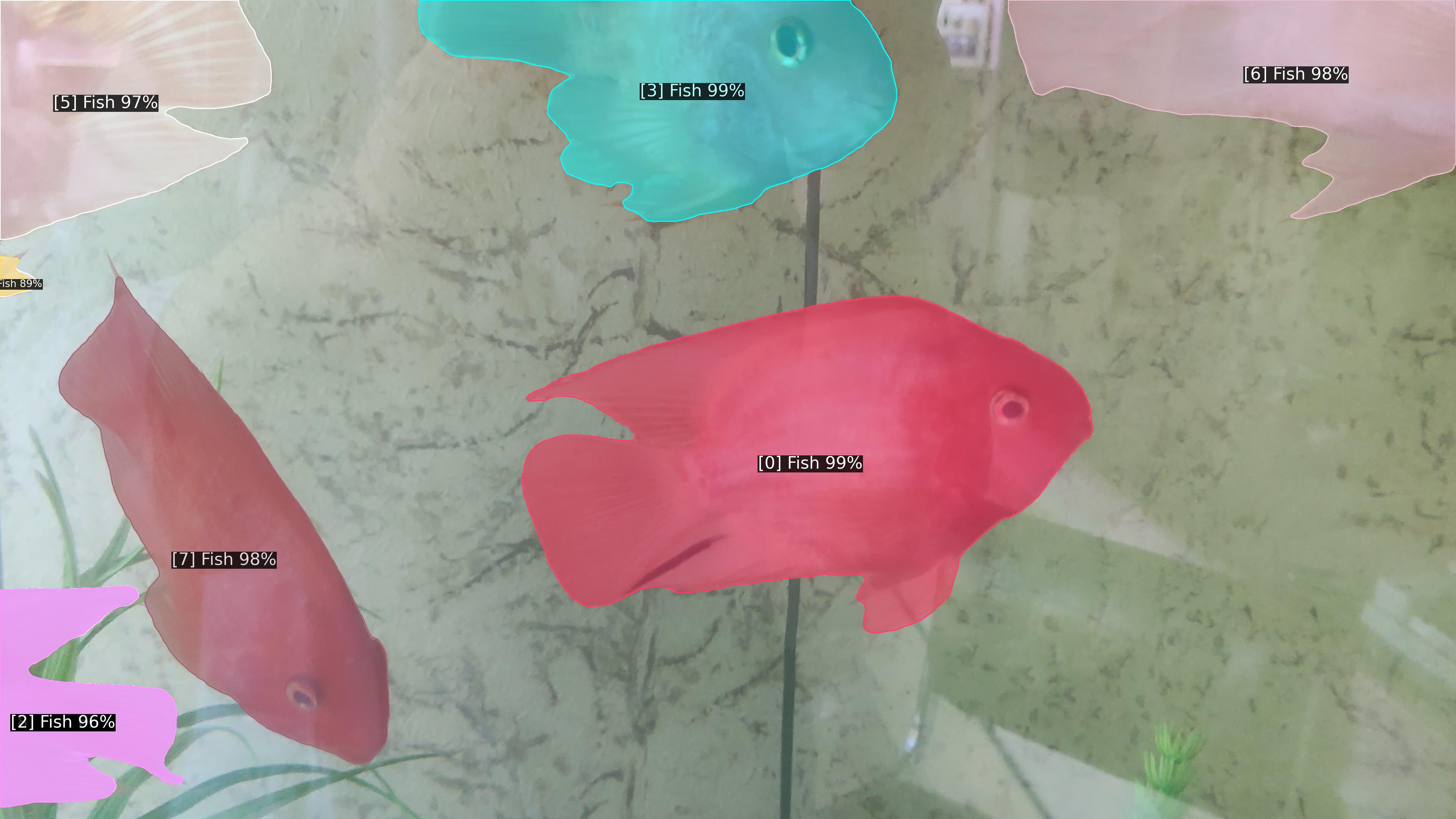} \\

\end{tabular}
\caption{More qualitative results of GRAtt-VIS on OVIS dataset (best viewed in color and zoom). Different colors refer to different tracks. The first video demonstrates the ability of \name{} to track  speeding objects in blurry video frames. The second one constitutes numerous instances of the same category compactly placed in a group. The third and fourth videos introduce occlusion of various degrees. The successful recovery from these challenging scenarios in lengthy videos proves our model's capability to hold on to \lq relevant' instance representations without any memory bank.}
\label{fig:qualitative}
\end{figure*}

\begin{figure*}[!h]
\centering
\begin{tabular}{cccc}
\includegraphics[width=.22\linewidth]{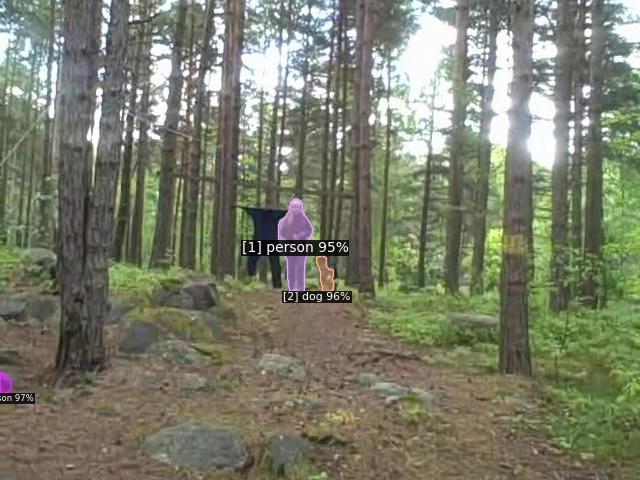} &
\includegraphics[width=.22\linewidth]{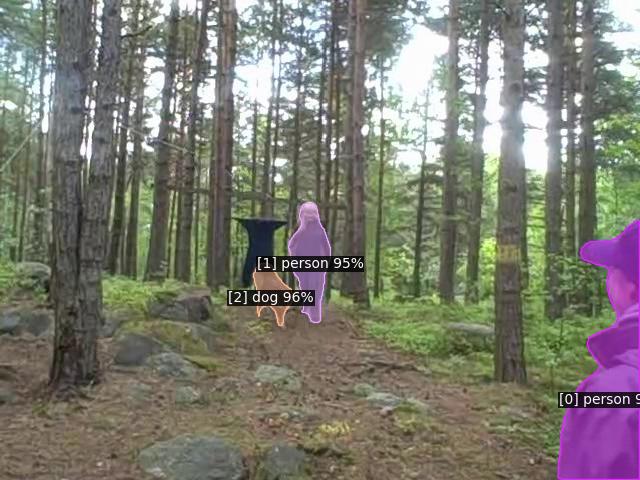} &
\includegraphics[width=.22\linewidth]{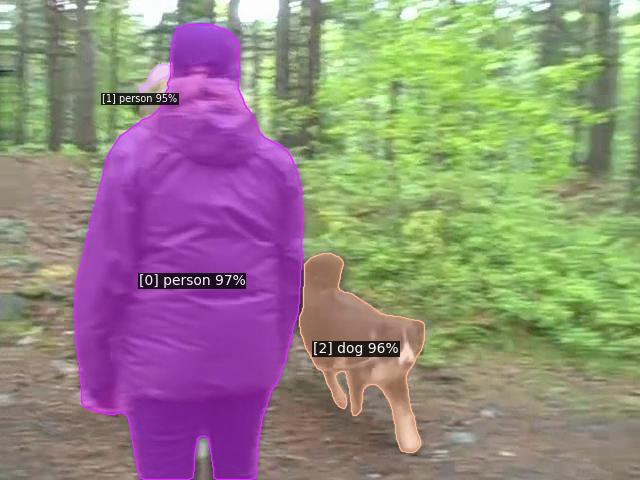} &
\includegraphics[width=.22\linewidth]{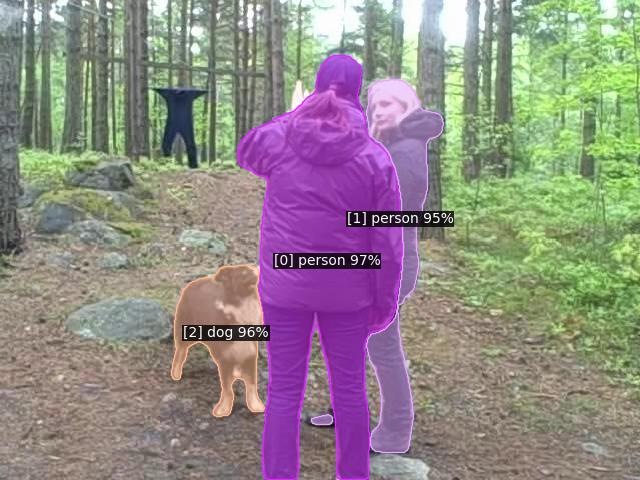} \\
\includegraphics[width=.22\linewidth]{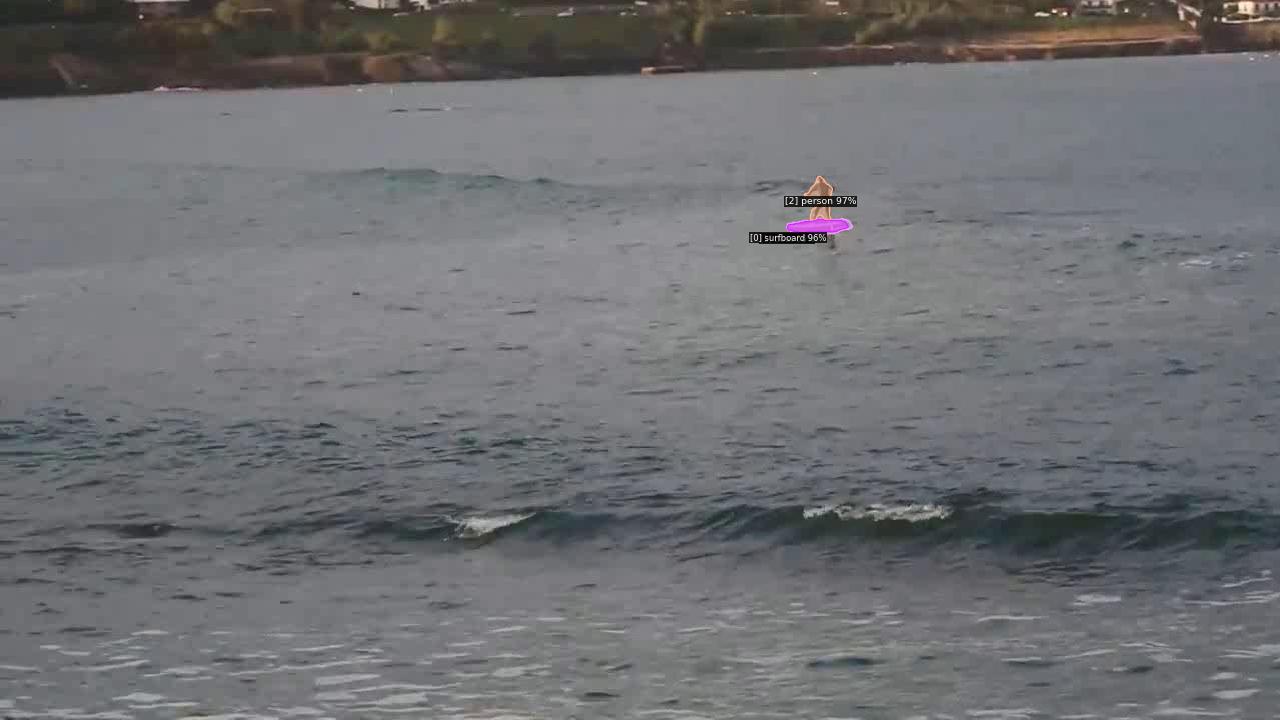} &
\includegraphics[width=.22\linewidth]{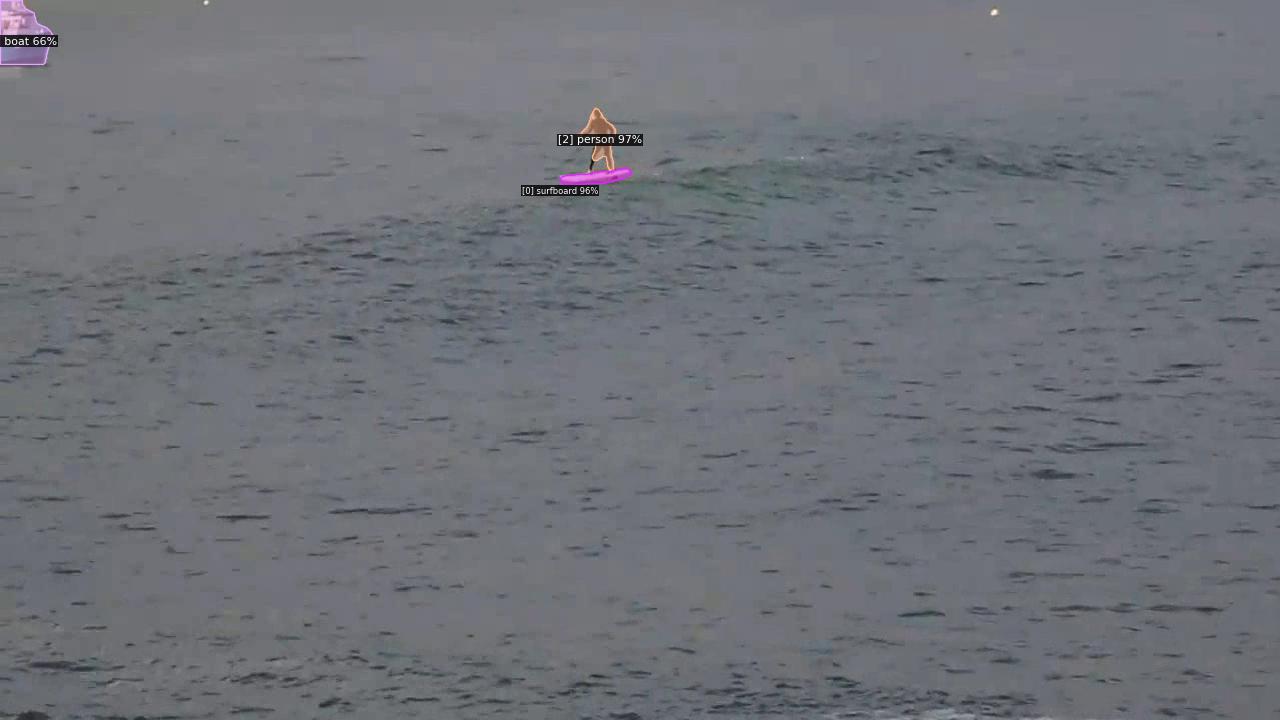} &
\includegraphics[width=.22\linewidth]{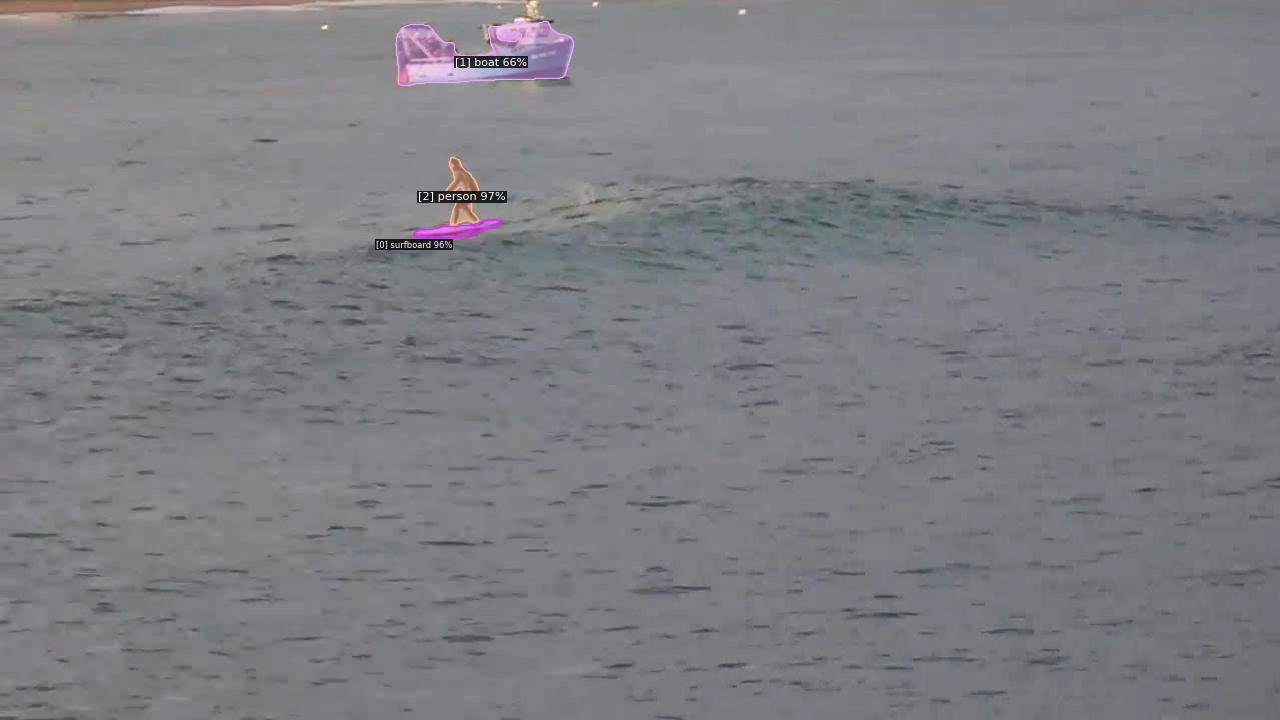} &
\includegraphics[width=.22\linewidth]{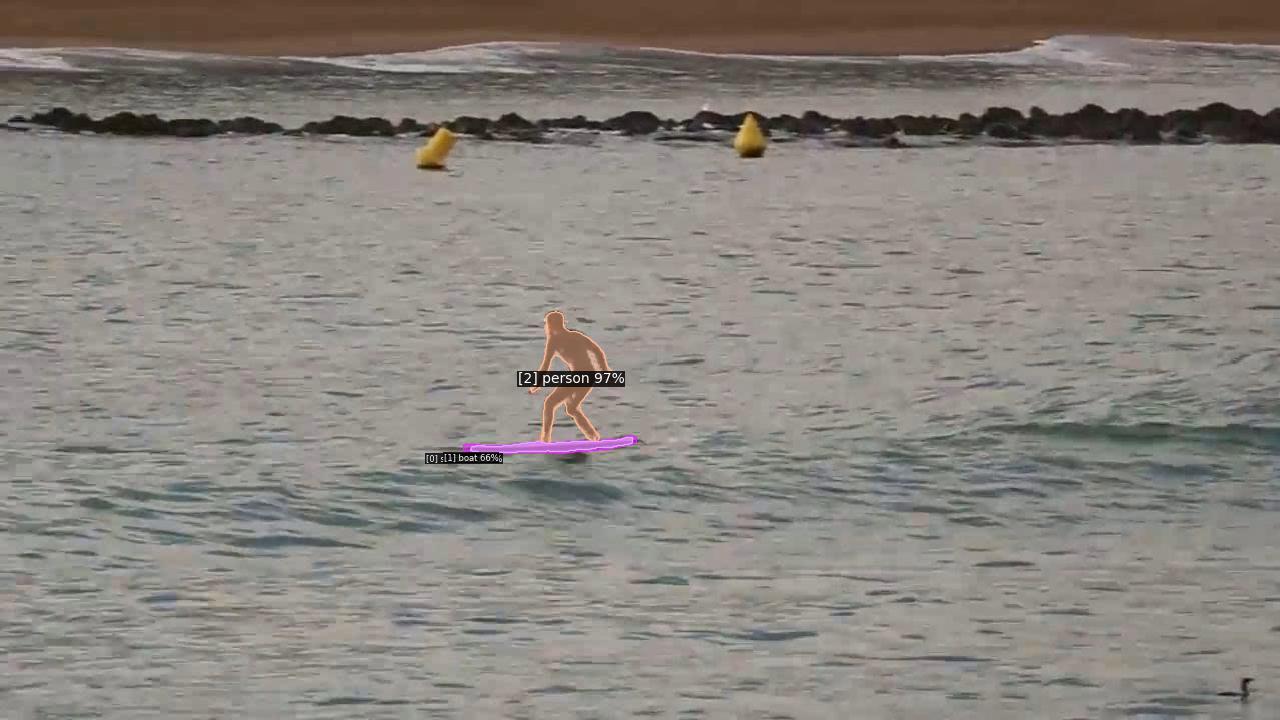} \\
\includegraphics[width=.22\linewidth]{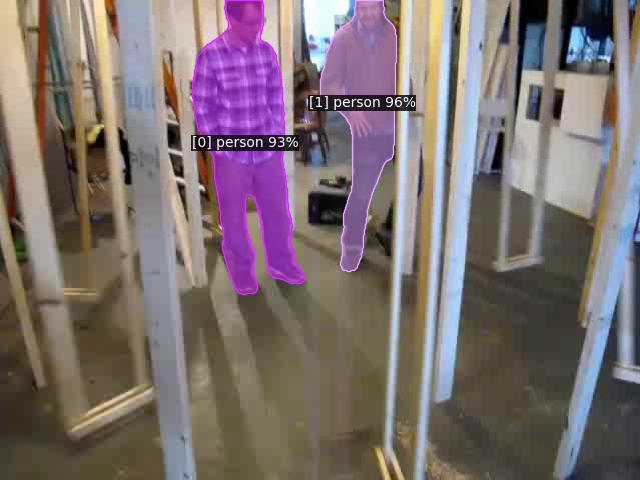} &
\includegraphics[width=.22\linewidth]{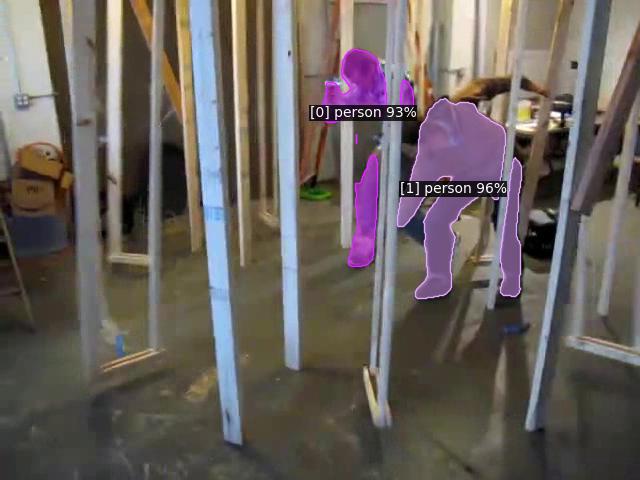} &
\includegraphics[width=.22\linewidth]{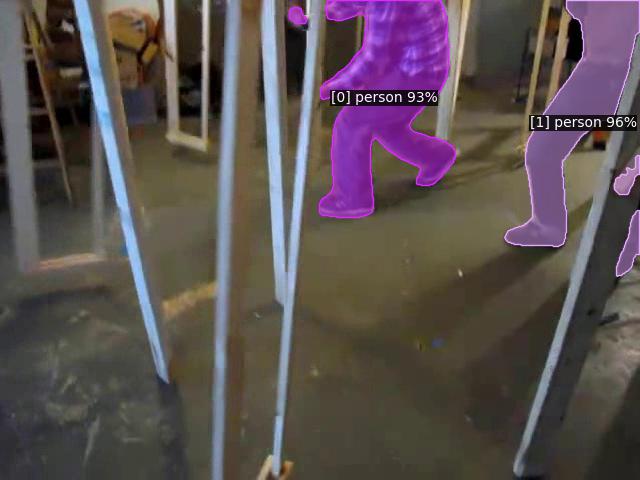} &
\includegraphics[width=.22\linewidth]{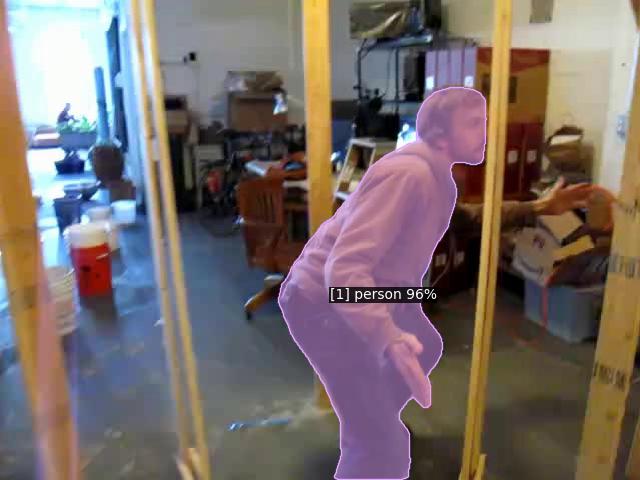} \\
\includegraphics[width=.22\linewidth]{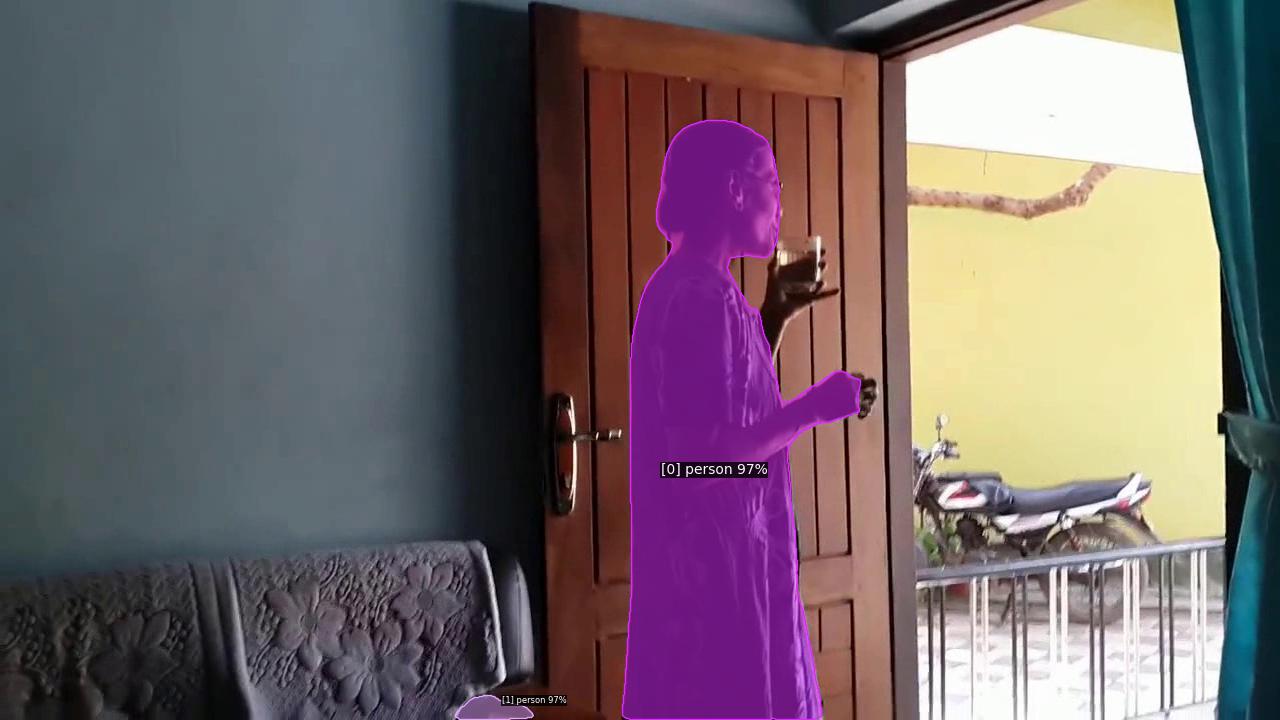} &
\includegraphics[width=.22\linewidth]{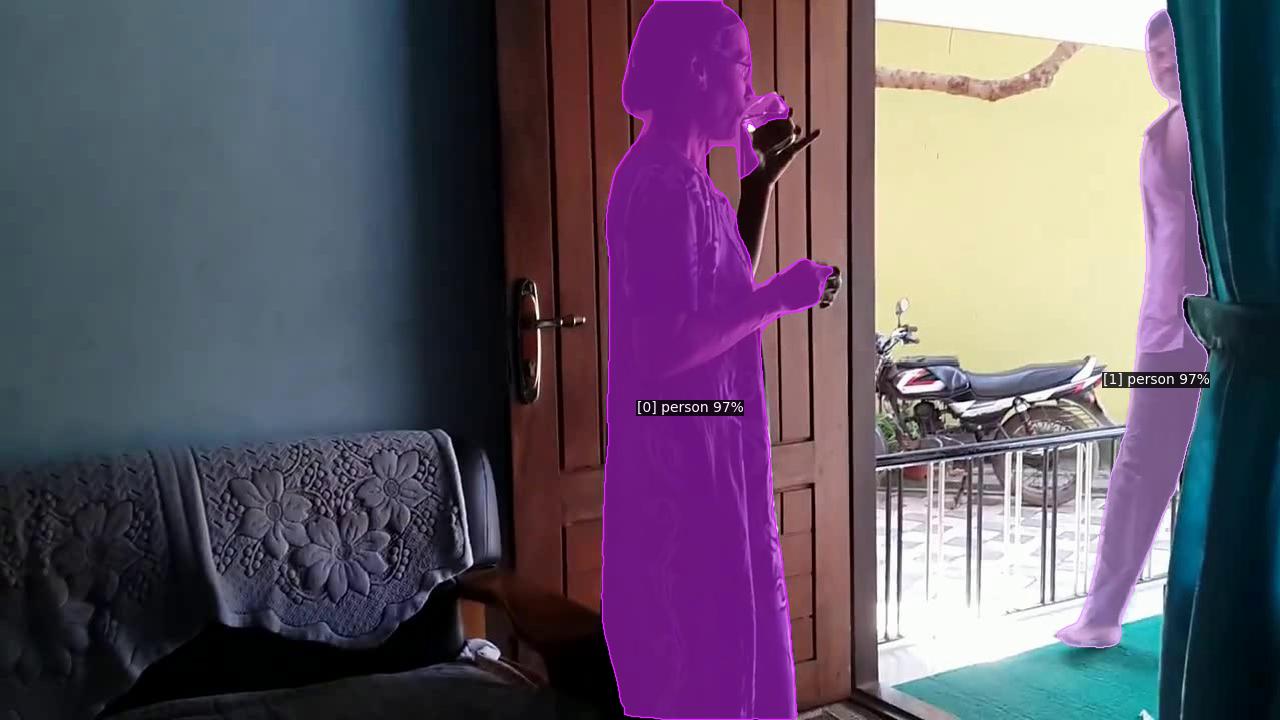} &
\includegraphics[width=.22\linewidth]{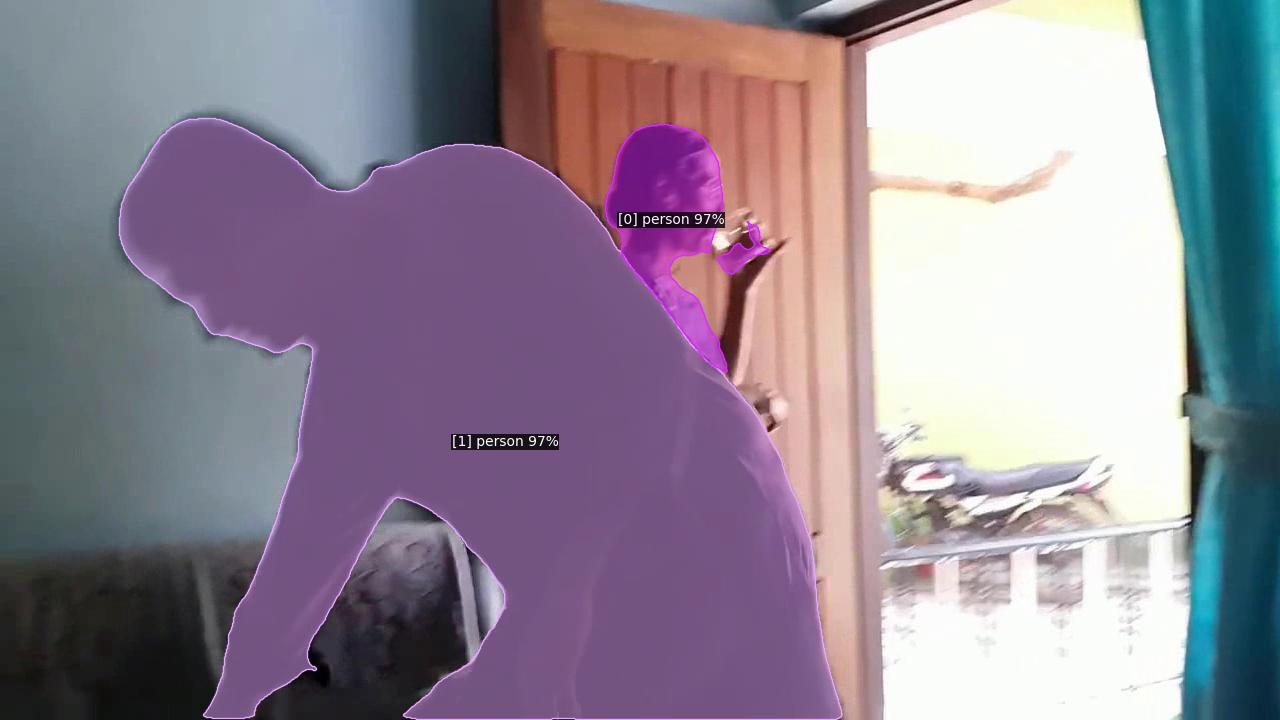} &
\includegraphics[width=.22\linewidth]{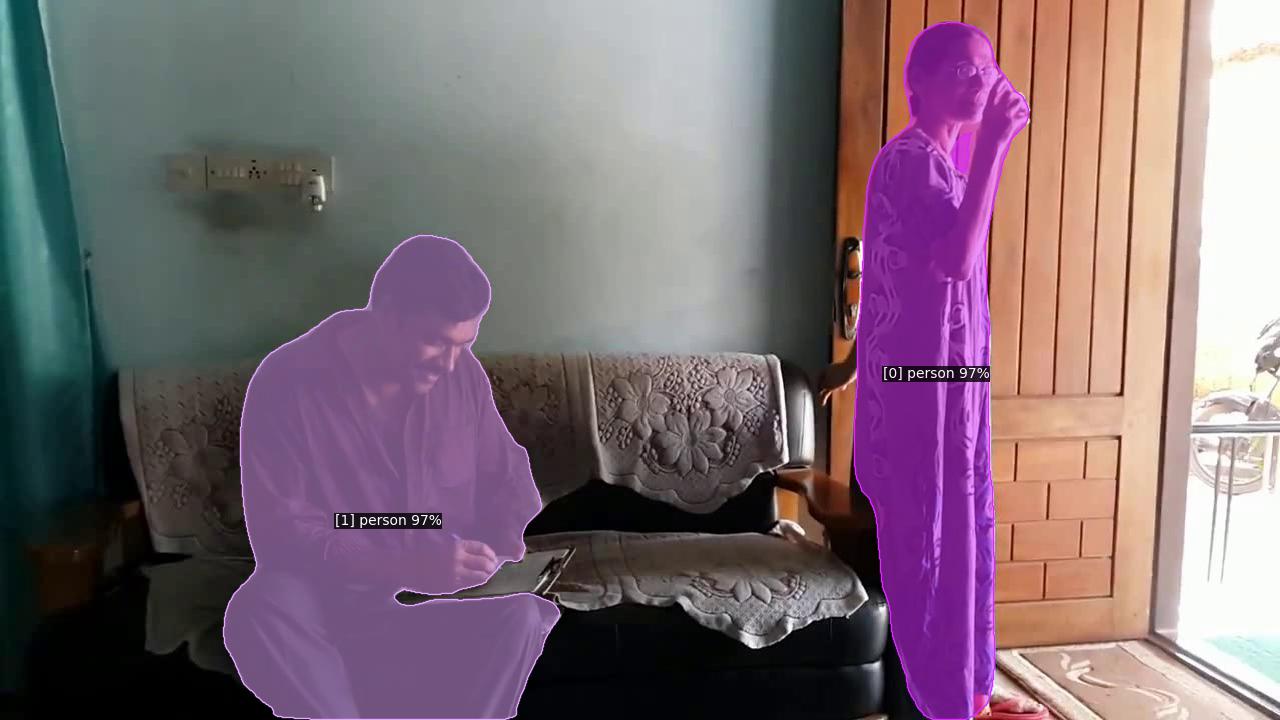} \\
\end{tabular}
\caption{Qualitative results of GRAtt-VIS on YTVIS-22 Long Video dataset. The first video has a new instance appearance. The second one also has object disappearance. The other two introduce complex camera motion and object disappearance.}
\label{fig:qualitative_yt22}
\end{figure*}

\end{document}